# Nonconvex Sparse Learning via Stochastic Optimization with Progressive Variance Reduction

Xingguo Li, Raman Arora, Han Liu, Jarvis Haupt, and Tuo Zhao*

**Abstract**

We propose a stochastic variance reduced optimization algorithm for solving large scale sparse learning problems with cardinality constraints. We provide sufficient conditions under which the proposed algorithm enjoys strong linear convergence guarantees and optimal estimation accuracy in high dimensions. We further extend the proposed algorithm to an asynchronous parallel variant with a near linear speedup. Numerical experiments demonstrate the efficiency of our algorithm in terms of both parameter estimation and computational performance.

## 1 Introduction

In machine learning problems, high dimensionality poses statistical as well as computational challenges. Researchers dealing with such problems often find refuge in the principle of parsimony – assuming that only a small number of variables are relevant for modeling the response variable, thereby making the analysis manageable and algorithmic design feasible. Consequently, in the past decade, a large family of $\ell_1$-regularized or $\ell_1$-constrained sparse estimators have been proposed, including Lasso (Tibshirani, 1996), Logistic Lasso (Van de Geer, 2008), Group Lasso (Yuan and Lin, 2006), Graphical Lasso (Banerjee et al., 2008; Friedman et al., 2008), and more. The $\ell_1$-norm serves as a convex surrogate for controlling the cardinality of the parameters, and a large family of algorithms, such as proximal gradient algorithms (Nesterov, 2013a), have been developed for finding the $\ell_1$-norm based estimators in polynomial time. The $\ell_1$-regularization or constraint, however, often incurs large estimation bias, and attains worse empirical performance than the $\ell_0$-regularization and constraint (Fan and Li, 2001; Zhang, 2010). This motivates us to study a family of cardinality constrained M-estimators. Formally, we consider the following nonconvex optimization problem:

$$\min_{\theta \in \mathbb{R}^d} \mathcal{F}(\theta) \quad \text{subject to } \|\theta\|_0 \leq k, \tag{1.1}$$

where $\mathcal{F}(\theta)$ is a potentially nonconvex loss function, and $\|\theta\|_0$ denotes the number of nonzero entries in $\theta$ (Yuan et al., 2014; Jain et al., 2014).

*Xingguo Li and Jarvis Haupt are affiliated with Department of Electrical and Computer Engineering at University of Minnesota Twin Cities; Raman Arora is affiliated with Department of Computer Science at Johns Hopkins University; Han Liu is affiliated with Department of Electrical Engineering and Computer Science and Statistics at Northwestern University; Tuo Zhao is affiliated with School of Industrial and Systems Engineering at Georgia Institute of Technology. Correspondence to: Xingguo Li <lixx1661@umn.edu> and Tuo Zhao <tourzhao@gatech.edu>. Some preliminary results in this paper were presented at the 33rd International Conference on Machine Learning (Li et al., 2016). This research is supported by NSF CCF-1217751; NSF AST-1247885; DARPA Young Faculty Award N66001-14-1-4047; NSF DMS-1454377-CAREER; NSF IIS-1546482-BIGDATA; NIH R01MH102339; NSF IIS-1408910; NSF IIS-1332109; NIH R01GM083084; Doctoral Dissertation Fellowship from University of Minnesota.

To solve (1.1), a (full) gradient hard thresholding (FG-HT) algorithm has been studied in statistics as well as machine learning over the past few years (Yuan et al., 2014; Jain et al., 2014; Blumensath and Davies, 2009; Foucart, 2011). FG-HT involves iteratively performing a gradient update followed by a hard thresholding operation – let $\mathcal{H}_k(\theta)$ denote the hard thresholding operator that keeps the largest $k$ entries in magnitude and sets the other entries equal to zero – then, at the $t$-th iteration, FG-HT performs the following update:

$$\theta^{(t)} = \mathcal{H}_k\Big(\theta^{(t-1)} - \eta \nabla \mathcal{F}(\theta^{(t-1)})\Big),$$

where $\nabla \mathcal{F}(\theta^{(t)})$ is the gradient of the objective at $\theta^{(t)}$ and $\eta > 0$ is a step size parameter. Existing literature has shown that under suitable conditions, FG-HT attains linear convergence to an approximately global optimum with optimal estimation accuracy, with high probability (Yuan et al., 2014; Jain et al., 2014).

Despite these good properties, FG-HT is not suitable for solving large-scale problems. The computational bottleneck stems from the fact that FG-HT evaluates the (full) gradient at each iteration; its computational complexity therefore depends linearly on the number of samples, making FG-HT computationally expensive for high-dimensional problems in large-scale settings.

To address the scalability issue, Nguyen et al. (2014) consider a scenario that is typical in machine learning wherein the objective function decomposes over samples, i.e. the objective function $\mathcal{F}(\theta)$ takes an additive form over many smooth component functions:

$$\mathcal{F}(\theta) = \frac{1}{n} \sum_{i=1}^{n} f_i(\theta) \quad \text{and} \quad \nabla \mathcal{F}(\theta) = \frac{1}{n} \sum_{i=1}^{n} \nabla f_i(\theta),$$

and each $f_i(\theta)$ is associated with a few samples of the entire data set (aka, the mini-batch setting). In such settings, we can exploit the additive nature of $\mathcal{F}(\theta)$ and consider a stochastic gradient hard thresholding (SG-HT) algorithm based on unbiased estimates of the gradient rather than computing the full gradient. In particular, the SG-HT algorithm uses a stochastic gradient $\nabla f_{i_t}(\theta^{(t)})$ as an estimate of the full gradient $\nabla \mathcal{F}(\theta^{(t)})$, where $i_t$ is sampled uniformly randomly from $\{1, \ldots, n\}$ at each iteration. Though SG-HT greatly reduces the computational cost at each iteration, it can only obtain an estimator with suboptimal estimation accuracy, owing to the variance of the stochastic gradient introduced by random sampling. Moreover, the convergence analysis of SG-HT (Nguyen et al., 2014) requires $\mathcal{F}(\theta)$ to satisfy the Restricted Isometry Property (RIP) with parameter $1/7$, i.e., the restricted condition number of the Hessian matrix $\nabla^2 \mathcal{F}(\theta)$ cannot exceed 4/3 (see more details in Section 3). Taking sparse linear regression as an example, such an RIP condition requires the design matrix to be nearly orthogonal, which is not satisfied even by some simple random correlated Gaussian designs (Raskutti et al., 2010).

To address the suboptimal estimation accuracy and the restrictive requirement on $\mathcal{F}(\theta)$ in the stochastic setting, we propose a stochastic variance reduced gradient hard thresholding (SVRG-HT) algorithm. More specifically, we propose a stochastic optimization scheme to reduce the variance introduced by the random sampling (Johnson and Zhang, 2013; Konečný and Richtárik, 2013). SVRG-HT contains two nested loops: at each iteration of the outer loop, SVRG-HT calculates the full gradient. In the subsequent inner loop, the stochastic gradient update is adjusted by the full gradient followed by hard thresholding. This simple modification enables the algorithm to attain linear convergence to an approximately global optimum with optimal estimation accuracy, and meanwhile the amortized computational complexity remains similar to that of conventional stochastic optimization. Moreover, our theoretical analysis is applicable to a large restricted condition number of the Hessian matrix $\nabla^2 \mathcal{F}(\theta)$, e.g., 100, rather than requires the restricted condition number to be a small fixed constant, e.g., < 2, as in Nguyen et al. (2014). To further boost the computational performance, we extend SVRG-HT to an asynchronous parallel variant via a lock-free approach for parallelization (Recht et al., 2011; Reddi et al., 2015; Liu et al., 2015). We establish theoretically that a near linear speedup is achieved for asynchronous SVRG-HT.

Several existing algorithms are closely related to our proposed algorithm, including the proximal stochastic variance reduced gradient algorithm (Xiao and Zhang, 2014), stochastic averaging gradient algorithm (Roux et al., 2012), and stochastic dual coordinate ascent algorithm (Shalev-Shwartz and Zhang, 2013). However, these algorithms guarantee global linear convergence only for strongly convex optimization problems. Several statistical methods in existing literature are also closely related to cardinality constrained M-estimators, including nonconvex constrained M-estimators (Shen et al., 2012) and nonconvex regularized M-estimators (Loh and Wainwright, 2013). These methods usually require somewhat complicated computational formulation and often involve many tuning parameters. We discuss these methods in more details in Section 6.

A preliminary conference version of this paper was published on the 33rd International Conference on Machine Learning (Li et al., 2016). After the conference version was accepted, we found an concurrent work released on arXiv (Shen and Li, 2016), which independently propose a similar algorithm to SVRG-HT. Here we make a clarification on the difference between their results and ours: (1) Shen and Li (2016) only consider the computational theory for sparse linear regression and sparse logistic regression, we develop computational as well as statistical theories not only for sparse linear regression, but also for generalized linear models and low-rank matrix estimation; (2) Our computational theory is sharper and more refined (See more details in Section 3.1); (3) We propose an asynchronous parallel extension with both computational and statistical guarantees.

The rest of the paper is organized as follows. We present the SVRG-HT algorithm in Section 2 and discuss the computational and statistical guarantees in Section 3. In Section 4, we introduce a parallel variant of SVRG-HT. Section 5 describes the numerical experiments. Related algorithms and optimization problems are discussed in Section 6. Proofs of all the technical results can be found in Section 7 along with technical details in the Appendix.

## 2 Algorithm

We first set up the requisite notation to present the proposed algorithm.

**Notation** Given an integer $n \geq 1$, we define $[n] = \{1,\ldots,n\}$. Given a vector $v = (v_1,\ldots,v_d)^\top \in \mathbb{R}^d$, we define vector norms: $\|v\|_1 = \sum_j |v_j|$, $\|v\|_2^2 = \sum_j v_j^2$, and $\|v\|_\infty = \max_j |v_j|$. Given an index set $\mathcal{I} \subseteq [d]$, we define $\mathcal{I}^C$ as the complement set of $\mathcal{I}$, and $v_\mathcal{I} \in \mathbb{R}^d$, where $[v_\mathcal{I}]_j = v_j$ if $j \in \mathcal{I}$ and $[v_\mathcal{I}]_j = 0$ if $j \notin \mathcal{I}$. We use $\mathrm{supp}(v)$ to denote the index set of nonzero entries of $v$. Given two vectors $v, w \in \mathbb{R}^d$, we use $\langle v, w \rangle = \sum_{i=1}^d v_i w_i$ to denote the inner product. Given a matrix $A \in \mathbb{R}^{n \times d}$, we use $A^\top$ to denote the transpose, $A_{i*}$ and $A_{*j}$ to denote the $i$-th row and $j$-th column respectively, $\sigma_i(A)$ to denote the $i$-th largest singular value, $\mathrm{rank}(A)$ to denote the rank, $\|A\|_* = \sum_{i=1}^{\mathrm{rank}(A)} \sigma_i(A)$ to denote the nuclear norm, and $\mathrm{vec}(A)$ to denote a vector obtained by concatenating the columns of $A$. Given an index set $\mathcal{I} \subseteq [d]$, we denote the submatrix of $A$ with all row indices in $\mathcal{I}$ by $A_{\mathcal{I}*}$, and denote the submatrix of $A$ with all column indices in $\mathcal{I}$ by $A_{*\mathcal{I}}$. Given two matrices $A, B \in \mathbb{R}^{n \times d}$, we use $\langle A, B \rangle = \mathrm{Trace}(A^\top B) = \sum_{i=1}^n \sum_{j=1}^d A_{ij} B_{ij}$. Moreover, we use the common notations of $\Omega(\cdot)$ and $\mathcal{O}(\cdot)$ to characterize the asymptotics of two real sequences. For logarithmic functions, we denote $\log(\cdot)$ as the natural logarithm when we do not specify the base.

**Algorithm** The proposed stochastic variance reduced gradient hard thresholding (SVRG-HT) algorithm is presented in Algorithm 1. In contrast to the stochastic gradient hard thresholding (SG-HT) algorithm (Nguyen et al., 2014), the proposed algorithm adopts a stochastic optimization scheme (Johnson and Zhang, 2013) that can guarantee that the variance introduced by stochastic sampling over component functions decreases with the optimization error. In order to illustrate the mechanics of SVRG-HT, we sketch a concrete example. Specifically, we consider a sparse linear model

$$y = A\theta^* + z, \tag{2.1}$$

where $A \in \mathbb{R}^{nb \times d}$ is the design matrix, $y \in \mathbb{R}^{nb}$ is the response vector, $\theta^* \in \mathbb{R}^d$ is the unknown sparse regression coefficient vector with $\|\theta^*\|_0 = k^*$, and $z \in \mathbb{R}^{nb}$ is a random noise vector sampled from $\mathcal{N}(0, \sigma^2 I)$. We are interested in estimating $\theta^*$ by sovling the following nonconvex optimization problem:

$$\min_{\theta \in \mathbb{R}^d} \mathcal{F}(\theta) = \frac{1}{2nb}\|y - A\theta\|_2^2 \quad \text{subject to } \|\theta\|_0 \leq k. \tag{2.2}$$

To solve (2.2) in the stochastic mini-batch optimization regime, we divide $A$ into $n$ submatrices such that each submatrix contains $b$ rows of $A$, i.e., we have $n$ mini-batches and $b$ is the mini-batch

**Algorithm 1** Stochastic Variance Reduced Gradient Hard Thresholding Algorithm (SVRG-HT). $\mathcal{H}_k(\cdot)$ is the hard thresholding operator, which keeps the largest $k$ (in magnitude) entries and sets the other entries equal to zero.

**Input:** update frequency $m$, step size parameter $\eta$, sparsity $k$, and initial solution $\widetilde{\theta}^{(0)}$
**for** $r = 1, 2, \ldots$ **do**
  $\widetilde{\theta} = \widetilde{\theta}^{(r-1)}$
  $\widetilde{\mu} = \frac{1}{n} \sum_{i=1}^{n} \nabla f_i(\widetilde{\theta})$
  $\theta^{(0)} = \widetilde{\theta}$
  **for** $t = 0, 1, \ldots, m-1$ **do**
    (S1) Randomly sample $i_t$ from $[n]$
    (S2) $\overline{\theta}^{(t+1)} = \theta^{(t)} - \eta \left( \nabla f_{i_t}(\theta^{(t)}) - \nabla f_{i_t}(\widetilde{\theta}) + \widetilde{\mu} \right)$
    (S3) $\theta^{(t+1)} = \mathcal{H}_k(\overline{\theta}^{(t+1)})$
  **end for**
  $\widetilde{\theta}^{(r)} = \theta^{(t+1)}$ for randomly chosen $t \in [m]$
**end for**

---

**Algorithm 2** Stochastic Average Gradient Hard Thresholding Algorithm (SAGA-HT). SAGA-HT has similar computational and statistical performance to SVRG-HT in both theory and practice.

**Input:** step size parameter $\eta$, sparsity $k$, and initial solution $\widetilde{\theta}^{(0)}$
**for** $r = 1, 2, \ldots$ **do**
  Randomly sample $i_r$ from $[n]$
  $\theta_{i_r}^{(r)} = \widetilde{\theta}^{(r-1)}$, and store $\nabla f_{i_t}(\theta_{i_t}^{(r)})$ in the table of stochastic gradients. All other entries
      in the table remain unchanged
  $\overline{\theta}^{(r)} = \widetilde{\theta}^{(r-1)} - \eta \left( \nabla f_{i_t}(\theta^{(t)}) - \nabla f_{i_t}(\theta_i^{(r-1)}) + \frac{1}{n} \sum_{i=1}^{n} \nabla f_i(\theta_i^{(r)}) \right)$
  $\widetilde{\theta}^{(r)} = \mathcal{H}_k(\overline{\theta}^{(t+1)})$
**end for**

---

size. For notational simplicity, we define the $i$-th submatrix as $A_{\mathcal{S}_i *}$, where $\mathcal{S}_i$ is the set of the corresponding row indices with $|\mathcal{S}_i| = b$ for all $i = 1, \ldots, n$. Accordingly, we have

$$f_i(\theta) = \frac{1}{2b} \|y_{\mathcal{S}_i} - A_{\mathcal{S}_i *} \theta\|_2^2 \quad \text{and} \quad \mathcal{F}(\theta) = \frac{1}{n} \sum_{i=1}^{n} \frac{1}{2b} \|y_{\mathcal{S}_i} - A_{\mathcal{S}_i *}\|_2^2.$$

Let us consider the computational cost of SVRG-HT per iteration. Note that the full gradient $\widetilde{\mu} = \nabla \mathcal{F}(\theta)$ remains unchanged through the inner loop, and we only calculate the full gradient once every $m$ inner iterations. We can verify that the average per iteration computational cost is $\mathcal{O}((n+m)bd/m)$. When $m$ is of the same order of $n$ for some constant $c > 0$, it is further reduced to $\mathcal{O}(bd)$, which matches that of SG-HT up to a constant factor.

A closely related algorithm to SVRG is stochastic average gradient algorithm (SAGA); we refer the reader to Defazio et al. (2014) for further details. In Algorithm 2, we present an extension of SAGA to SAGA hard thresholding (SAGA-HT) algorithm for nonconvex sparse learning. As for SVRG-HT, the average per iteration computational cost for SAGA-HT is $\mathcal{O}(bd)$. However, unlike SAGA-HT, which needs to maintain $n$ stochastic gradients in the memory resulting in a space complexity of $\mathcal{O}(nd)$, SVRG-HT only maintains a batch gradient in memory relaxing the space requirements to $\mathcal{O}(d)$. This is an enormous advantage for SVRG-HT over SAGA-HT for large $n$.

## 3 Theory

We are interested in analyzing the convergence of our proposed algorithm to the unknown sparse parameter $\theta^*$ of the underlying statistical model. For example, for sparse linear regression in (2.1), $\theta^*$ is the unknown regression coefficient vector. This is different from the conventional optimization theory, which analyzes the convergence properties of the algorithm to an optimum of the optimization problem.

Our proposed theoretical analysis is applicable to both SVRG-HT and SAGA-HT. As mentioned in Section 2, SVRG-HT has an advantage over SAGA-HT in space complexity. Therefore, we focus only on the analysis for SVRG-HT in this section, and an extension to SAGA-HT is straightforward.

Throughout the analysis, we make two important assumptions on the objective function, which are defined as follows.

**Definition 3.1** (Restricted Strong Convexity Condition). A differentiable function $\mathcal{F}$ is restricted $\rho_s^-$-strongly convex at sparsity level $s$ if there exists a generic constant $\rho_s^- > 0$ such that for any $\theta, \theta' \in \mathbb{R}^d$ with $\|\theta - \theta'\|_0 \leq s$, we have

$$\mathcal{F}(\theta) - \mathcal{F}(\theta') - \langle \nabla \mathcal{F}(\theta'), \theta - \theta' \rangle \geq \frac{\rho_s^-}{2} \|\theta - \theta'\|_2^2. \tag{3.1}$$

**Definition 3.2** (Restricted Strong Smoothness Condition). For any $i \in [n]$, a differentiable function $f_i$ is restricted $\rho_s^+$-strongly smooth at sparsity level $s$ if there exists a generic constant $\rho_s^+ > 0$ such that for any $\theta, \theta' \in \mathbb{R}^d$ with $\|\theta - \theta'\|_0 \leq s$, we have

$$f_i(\theta) - f_i(\theta') - \langle \nabla f_i(\theta'), \theta - \theta' \rangle \leq \frac{\rho_s^+}{2} \|\theta - \theta'\|_2^2. \tag{3.2}$$

We assume that the objective function $\mathcal{F}(\theta)$ satisfies the restricted strong convexity (RSC) condition, and all component functions $\{f_i(\theta)\}_{i=1}^n$ satisfy the restricted strong smoothness (RSS) condition. Moreover, we define the restricted condition number $\kappa_s = \rho_s^+/\rho_s^-$. RSC and RSS conditions have been widely studied in high dimensional statistical theory (Raskutti et al., 2010; Loh and Wainwright, 2013; Agarwal et al., 2010). They guarantee that the objective function behaves like a strongly convex and smooth function over a sparse domain even the function is nonconvex. An example of nonconvex $\mathcal{F}$ is provided in Remark 3.12. RSC and RSS conditions are extremely important for establishing the computational theory.

The restricted isometry property (RIP) is closely related to the RSC and RSS conditions (Candes et al., 2006; Candes and Plan, 2011). However, RIP is more restrictive, since it requires $\rho_s^+ < 2$, which can be easily violated by simple random correlated sub-Gaussian designs. Moreover, RIP is only applicable to linear regression, while the RSC and RSS conditions are applicable to more general problems such as sparse generalized linear models estimation.

## 3.1 Computational Theory

We present two key technical lemmas that will be instrumental in developing computational theory for SVRG-HT. Recall that $\theta^* \in \mathbb{R}^d$ is the unknown sparse vector of interest with $\|\theta^*\|_0 \le k^*$, and $\mathcal{H}_k(\cdot): \mathbb{R}^d \to \mathbb{R}^d$ is a hard thresholding operator that keeps the largest $k$ entries (in magnitude) setting other entries to zero.

**Lemma 3.3.** For $k > k^*$ and for any vector $\theta \in \mathbb{R}^d$, we have

$$\|\mathcal{H}_k(\theta) - \theta^*\|_2^2 \le \left(1 + \frac{2\sqrt{k^*}}{\sqrt{k - k^*}}\right) \|\theta - \theta^*\|_2^2. \tag{3.3}$$

Lemma 3.3 shows that the hard thresholding operator is nearly non-expansive for $k$ sufficiently larger than $k^*$ such that $\frac{2\sqrt{k^*}}{\sqrt{k-k^*}}$ is small. The proof of Lemma 3.3 is presented in Appendix 9.2.

**Remark 3.4.** It is important to note that while Lemma 3.3 may seem related to Lemma 1 in Jain et al. (2014), there is an important difference. Lemma 1 in Jain et al. (2014) characterizes the effect of the hard thresholding operator by bounding the distance $\|\mathcal{H}_k(\theta) - \theta\|_2$ between a vector and its thresholded version. Lemma 3.3, on the other hand, bounds the increase in distance of a vector from a fixed target vector (of sparsity $k^*$) due to thresholding. The latter, we argue, makes more intuitive sense from an optimization perspective.

For notational simplicity, we denote the full gradient and the stochastic variance reduced gradient by

$$\widetilde{\mu} = \nabla \mathcal{F}(\widetilde{\theta}) = \frac{1}{n} \sum_{i=1}^{n} \nabla f_i(\widetilde{\theta}) \quad \text{and} \quad g^{(t)}(\theta^{(t)}) = \nabla f_{i_t}(\theta^{(t)}) - \nabla f_{i_t}(\widetilde{\theta}) + \widetilde{\mu}. \tag{3.4}$$

The next lemma shows that $g^{(t)}(\theta^{(t-1)})$ is an unbiased estimator of $\nabla \mathcal{F}(\theta^{(t-1)})$ with a well controlled second moment over a sparse support.

**Lemma 3.5.** Suppose that $\mathcal{F}(\theta)$ satisfies the RSC condition and that functions $\{f_i(\theta)\}_{i=1}^n$ satisfy the RSS condition with $s = 2k + k^*$. Let $\mathcal{I}^* = \mathrm{supp}(\theta^*)$ denote the support of $\theta^*$. Let $\theta^{(t)}$ be a sparse vector with $\|\theta^{(t)}\|_0 \le k$ and support $\mathcal{I}^{(t)} = \mathrm{supp}(\theta^{(t)})$. Then conditioning on $\theta^{(t)}$, for any $\mathcal{I} \supseteq (\mathcal{I}^* \cup \mathcal{I}^{(t)})$, we have $\mathbb{E}[g^{(t)}(\theta^{(t)})] = \nabla \mathcal{F}(\theta^{(t)})$ and

$$\mathbb{E}\|g_\mathcal{I}^{(t)}(\theta^{(t)})\|_2^2 \le 12\rho_s^+ \left[\mathcal{F}(\theta^{(t)}) - \mathcal{F}(\theta^*) + \mathcal{F}(\widetilde{\theta}) - \mathcal{F}(\theta^*)\right] + 3\|\nabla_\mathcal{I} \mathcal{F}(\theta^*)\|_2^2. \tag{3.5}$$

The proof of Lemma 3.5 is presented in Appendix 9.3.

**Remark 3.6.** For smooth convex problems, we have $\nabla \mathcal{F}(\theta^*) = 0$ if $\theta^*$ is a global minimizer. However, given that the problem of interest here, Problem 1.1, is nonconvex, the second term on the R.H.S of (3.5) is nonzero. This results in a setting different from existing work using variance reduction (Johnson and Zhang, 2013; Konečný and Richtárik, 2013).

We now present our first main result characterizing the quality of solution given by Algorithm 1, both in terms of the error in the objective value as well as error in terms of the parameter estimation.

**Theorem 3.7.** Let $\theta^*$ denote the unknown sparse parameter vector of the underlying statistical model, with $\|\theta^*\|_0 \leq k^*$. Assume that the objective function $\mathcal{F}(\theta)$ satisfies the RSC condition and functions $\{f_i(\theta)\}_{i=1}^n$ satisfy the RSS condition with $s = 2k + k^*$, where $k \geq C_1 \kappa_s^2 k^*$ and $C_1$ is a generic constant. Define

$$\widetilde{\mathcal{I}} = \mathrm{supp}\left(\mathcal{H}_{2k}(\nabla \mathcal{F}(\theta^*))\right) \cup \mathrm{supp}(\theta^*).$$

Then, there exist generic constants $C_2, C_3$, and $C_4$ such that by setting $\eta \rho_s^+ \in [C_2, C_3]$ and $m \geq C_4 \kappa_s$, we have

$$\frac{\left(1 + \frac{2\sqrt{k^*}}{\sqrt{k-k^*}}\right)^m \cdot \frac{2\sqrt{k^*}}{\sqrt{k-k^*}}}{\eta \rho_s^- (1 - 6\eta\rho_s^+)\left(\left(1 + \frac{2\sqrt{k^*}}{\sqrt{k-k^*}}\right)^m - 1\right)} + \frac{6\eta\rho_s^+}{1 - 6\eta\rho_s^+} \leq \frac{3}{4}. \quad (3.6)$$

Furthermore, the parameter $\widetilde{\theta}^{(r)}$ at the $r$-th iteration of SVRG-HT satisfies

$$\mathbb{E}\left[\mathcal{F}(\widetilde{\theta}^{(r)}) - \mathcal{F}(\theta^*)\right] \leq \left(\frac{3}{4}\right)^r \cdot \left[\mathcal{F}(\widetilde{\theta}^{(0)}) - \mathcal{F}(\theta^*)\right] + \frac{6\eta}{(1-6\eta\rho_s^+)}\|\nabla_{\widetilde{\mathcal{I}}}\mathcal{F}(\theta^*)\|_2^2$$
$$+ \sqrt{s}\|\nabla\mathcal{F}(\theta^*)\|_\infty \mathbb{E}\|\widetilde{\theta}^{(r-1)} - \theta^*\|_2 \quad \text{and} \quad (3.7)$$

$$\mathbb{E}\|\widetilde{\theta}^{(r)} - \theta^*\|_2 \leq \sqrt{\frac{2\left(\frac{3}{4}\right)^r\left[\mathcal{F}(\widetilde{\theta}^{(0)}) - \mathcal{F}(\theta^*)\right]}{\rho_s^-}} + \frac{\sqrt{s}\|\nabla\mathcal{F}(\theta^*)\|_\infty}{\rho_s^-} + \|\nabla_{\widetilde{\mathcal{I}}}\mathcal{F}(\theta^*)\|_2\sqrt{\frac{12\eta}{(1-6\eta\rho_s^+)\rho_s^-}}$$
$$+ \sqrt{\frac{2\sqrt{s}\|\nabla\mathcal{F}(\theta^*)\|_\infty \mathbb{E}\|\widetilde{\theta}^{(r-1)} - \theta^*\|_2}{\rho_s^-}}. \quad (3.8)$$

Moreover, given a constant $\delta \in (0,1)$ and a pre-specified accuracy $0 < \varepsilon < \frac{2\sqrt{s}\|\nabla\mathcal{F}(\theta^*)\|_\infty}{\rho_s^-}$, we need at most

$$r = \left\lceil 4\log\left(\frac{\mathcal{F}(\widetilde{\theta}^{(0)}) - \mathcal{F}(\theta^*)}{\varepsilon\delta}\right) + 2\log\frac{\left|\|\widetilde{\theta}^{(0)} - \theta^*\|_2 - \frac{7\sqrt{s}\|\nabla\mathcal{F}(\theta^*)\|_\infty}{\rho_s^-}\right|}{\varepsilon\delta}\right\rceil \quad (3.9)$$

outer iterations to guarantee that with probability at least $1 - 2\delta$, we have

$$\mathcal{F}(\widetilde{\theta}^{(r)}) - \mathcal{F}(\theta^*) \leq \varepsilon\left(1 + \sqrt{s}\|\nabla\mathcal{F}(\theta^*)\|_\infty\right) + \frac{6\eta}{(1-6\eta\rho_s^+)}\|\nabla_{\widetilde{\mathcal{I}}}\mathcal{F}(\theta^*)\|_2^2 \quad \text{and} \quad (3.10)$$

$$\|\widetilde{\theta}^{(r)} - \theta^*\|_2 \leq \sqrt{\frac{2\varepsilon\left(1 + \sqrt{s}\|\nabla\mathcal{F}(\theta^*)\|_\infty\right)}{\rho_s^-}} + \frac{\sqrt{s}\|\nabla\mathcal{F}(\theta^*)\|_\infty}{\rho_s^-} + \|\nabla_{\widetilde{\mathcal{I}}}\mathcal{F}(\theta^*)\|_2\sqrt{\frac{12\eta}{(1-6\eta\rho_s^+)\rho_s^-}}. \quad (3.11)$$

The proof of Theorem 3.7 is presented in Section 7.1.

**Remark 3.8.** Theorem 3.7 has two important implications: **(I)** our analysis for SVRG-HT allows an arbitrary large $\kappa_s$ as long as $\mathcal{F}(\theta)$ and $\{f_i(\theta)\}_{i=1}^n$ satisfy the RSC and RSS conditions respectively with $s = \Omega(\kappa_s^2 k^*)$. In contrast, the theoretical analysis for SG-HT in Nguyen et al. (2014) requires $\kappa_s$ not to exceed $4/3$, which is very restrictive; **(II)** to get $\widetilde{\theta}^{(r)}$ to satisfy (3.10) and (3.11), we need $\mathcal{O}(\log(1/\varepsilon))$ outer iterations. Since within each outer iteration, we need to calculate a full gradient and $m$ stochastic variance reduced gradients, the overall computational complexity of SVRG-HT is

$$\mathcal{O}\left([nb + \kappa_s] \cdot \log\left(\frac{1}{\varepsilon}\right)\right),$$

where $b$ is the mini-batch size for each $f_i$. In contrast, the overall computational complexity of the full gradient hard thresholding algorithm (FG-HT) is $\mathcal{O}(\kappa_s nb \log(1/\varepsilon))$. Thus SVRG-HT yields a significant improvement over FG-HT when $\kappa_s$ is large.

## 3.2 Statistical Theory

SVRG-HT is applicable to a large family of sparse learning problems. Here, we present theoretical results for three popular examples of constrained M-estimation problems: sparse linear regression, sparse generalized linear model estimation, and low-rank matrix estimation (where the cardinality constraint is replaced by a rank constraint).

### 3.2.1 Sparse Linear Regression

Consider the sparse linear model

$$y = A\theta^* + z,$$

as introduced in Section 2. We want to estimate $\theta^*$ by solving the optimization problem in (2.2). We assume that for any $v \in \mathbb{R}^d$ with $\|v\|_0 \leq s$, the design matrix $A$ satisfies the *restricted eigenvalue* (RE) conditions:

$$\frac{\|Av\|_2^2}{nb} \geq \psi_1 \|v\|_2^2 - \varphi_1 \frac{\log d}{nb} \|v\|_1^2 \quad \text{and} \quad \frac{\|A_{\mathcal{S}_{i^*}} v\|_2^2}{b} \leq \psi_2 \|v\|_2^2 + \varphi_2 \frac{\log d}{b} \|v\|_1^2, \forall i \in [n], \qquad (3.12)$$

where $\psi_1$, $\psi_2$, $\varphi_1$, and $\varphi_2$ are constants that do not scale with $(n, b, k^*, d)$. Existing literature has shown that (3.12) is satisfied by many common examples of sub-Gaussian random design (Raskutti et al., 2010; Agarwal et al., 2010). The next lemma shows that (3.12) implies the RSC and RSS conditions.

**Lemma 3.9.** Suppose that the design matrix $A$ satisfies (3.12). Then, given large enough $n$ and $b$, there exist a constant $C_5$ and an integer $k$ such that $\mathcal{F}(\theta)$ and $\{f_i(\theta)\}_{i=1}^n$ satisfy the RSC and RSS conditions respectively with $s = 2k + k^*$, where

$$k = C_5 k^* \geq C_1 \kappa_s^2 k^*, \quad \rho_s^- \geq \psi_1/2, \quad \text{and} \quad \rho_s^+ \leq 2\psi_2.$$

A proof of Lemma 3.9 can be found in Appendix 9.4. Combining Lemma 3.9 and Theorem 3.7, we get the following computational and statistical guarantees for the estimator obtained by SVRG-HT.

**Corollary 3.10.** Suppose that the design matrix $A$ satisfies the RE conditions (3.12) with $\frac{\max_j \|A_{*j}\|_2}{\sqrt{nb}} \leq 1$, and $k$, $\eta$, and $m$ are as specified in Theorem 3.7. Then, for any confidence parameter $\delta \in (0,1)$, a sufficiently small accuracy parameter $\varepsilon > 0$, and large enough $n$ and $b$, we need at most $r$ outer iterations given in (3.9) via SVRG-HT to guarantee that with high probability, we have

$$\|\widetilde{\theta}^{(r)} - \theta^*\|_2 = \mathcal{O}\left(\sigma \sqrt{\frac{k^* \log d}{nb}}\right). \tag{3.13}$$

See Section 7.2 for a proof of Corollary 3.10.

**Remark 3.11.** Corollary 3.10 guarantees that the proposed SVRG-HT estimator attains the optimal statistical rate of convergence in parameter estimation (Raskutti et al., 2011) when $\varepsilon = \mathcal{O}\left(\sigma \sqrt{\frac{k^* \log d}{nb}}\right)$. In contrast, previous work, for instance see Corollary 5 in Nguyen et al. (2014), shows that the estimator obtained by the SGHT algorithm attains the statistical rate of convergence

$$\mathcal{O}\left(\sigma \sqrt{\frac{k^* \log d}{b}}\right)$$

with high probability, and hence is suboptimal when $n$ scales with $(b, k^*, d)$. This is because our estimation error depends on the full gradient $\|\nabla_{\widetilde{\mathcal{I}}} \mathcal{F}(\theta^*)\|_2$, but the estimation error in SGHT depends on the stochastic gradient, e.g., $\mathbb{E}_i \max \|\nabla_{\widetilde{\mathcal{I}}} f_i(\theta^*)\|_2$ in Nguyen et al. (2014), which is associated with the variance of $\|\nabla_{\widetilde{\mathcal{I}}} f_i(w^*)\|_2$ that is larger than the variance of $\|\nabla_{\widetilde{\mathcal{I}}} \mathcal{F}(\theta^*)\|_2$.

**Remark 3.12** (Nonconvex Objective). The objective function $\mathcal{F}$ can be nonconvex in general. We exemplify nonconvex $\mathcal{F}$ for the sparse linear regression following Loh and Wainwright (2012), when $A$ has additive noise, missing entries, or multiplicative noise. In specific, we consider the optimization problem:

$$\min_\theta \mathcal{F}(\theta) = \frac{1}{2} \theta^\top \widehat{\Gamma} \theta - \widehat{b}^\top \theta \quad \text{subject to } \|\theta\|_0 \leq k,$$

where $\widehat{\Gamma}$ and $\widehat{b}$ will be specified for each case accordingly in the following. In high dimensions, where $d \gg nb$, we will specify the case when $\widehat{\Gamma}$ is not positive semi-definite (PSD), which results in nonconvex $\mathcal{F}$.

We provide further details of the missing data scenario. Suppose that given a real value $\rho \in (0,1)$, we observe

$$Z_{ij} = \begin{cases} A_{ij}, & \text{with probability } 1 - \rho, \\ 0, & \text{otherwise.} \end{cases}$$

Then we set

$$\widehat{\Gamma} = \frac{\widetilde{Z}^\top \widetilde{Z}}{nb} - \rho \cdot \text{diag}\left(\frac{\widetilde{Z}^\top \widetilde{Z}}{nb}\right), \ \widehat{b} = \frac{\widetilde{Z}^\top y}{nb} \ \text{ and } \ \widetilde{Z}_{ij} = Z_{ij}/\rho.$$

For $\rho > 0$, since $\frac{\widetilde{Z}^\top \widetilde{Z}}{nb}$ has rank at most $nb$, the subtraction of a diagonal matrix may cause $\widehat{\Gamma}$ to have negative eigenvalues when $d \gg nb$. Thus, $\widehat{\Gamma}$ is not PSD.

Both computational and statistical results (Theorem 3.7 and Corollary 3.10) still hold for nonconvex $\mathcal{F}$ discussed above under the RE conditions, combining the analysis in Loh and Wainwright (2012). Further discussion for nonconvex $\mathcal{F}$ is provided in Appendix 9.1 when $A$ has additive noise or multiplicative noise. We also remark that a similar approach is proposed to accommodate heavy-tailed noise and design in sparse linear regression in high dimensions (Babacan et al., 2012), where the resulting $\mathcal{F}$ can be nonconvex.

### 3.2.2 Sparse Generalized Linear Models

We next consider sparse generalized linear models (GLM) defined by the following conditional distribution

$$\mathbb{P}(y_i | A_{i*}, \theta^*, \sigma) = \exp\left\{\frac{y_i A_{i*}\theta^* - h(A_{i*}\theta^*)}{a(\sigma)}\right\},$$

where $a(\sigma)$ is a fixed and known scale parameter, $\theta^* \in \mathbb{R}^d$ is the unknown sparse regression coefficient with $\|\theta^*\|_0 = k^*$, and $h(\cdot)$ is the cumulant function (Lehmann and Casella, 2006). For exponential families, the first derivative of the cumulant function satisfies

$$h'(A_{i*}\theta^*) = \mathbb{E}[y_i | A_{i*}, \theta^*, \sigma].$$

We further assume that the second derivative of the cumulant function is bounded, i.e. there exists some constant $c_u$ such that $h''(x) \leq c_u$ for all $x \in \mathbb{R}$. Such a boundedness assumption is necessary to establish the RSC and RSS conditions for GLM (Loh and Wainwright, 2013). Note that this assumption holds for various popular settings, including linear regression, logistic regression, and multinomial regression.

Analogous to sparse linear regression, we divide $A$ into $n$ mini-batches, where each mini-batch is denoted by $A_{\mathcal{S}_i *}$ and $\mathcal{S}_i$ denotes the corresponding row indices of $A$, with $|\mathcal{S}_i| = b$, for all $i = 1, ..., n$. Then, our objective is essentially the negative log-likelihood, i.e.,

$$\min_{\theta \in \mathbb{R}^d} \mathcal{F}(\theta) = \frac{1}{a(\sigma) \cdot n} \sum_{i=1}^{n} f_i(\theta) \text{ subject to } \|\theta\|_0 \leq k, \|\theta\|_2 \leq \tau, \tag{3.14}$$

for some $\tau > 0$, where $f_i(\theta) = \frac{1}{b}\sum_{\ell \in \mathcal{S}_i}(h(A_{\ell *}\theta) - y_\ell A_{\ell *}\theta)$, for all $i = 1, ..., n$. The additional constraint $\|\theta\|_2 \leq \tau$ in (3.14) may not be necessary in practice, but it is essential for our theoretical analysis; we further expand on this later in this section.

For concreteness, we consider sparse logistic regression as a special case of the setup above. We want to estimate $\theta^*$ from $nb$ independent responses $y_\ell \sim \text{Bernoulli}(\pi_\ell(\theta^*))$, $\ell \in [nb]$, where $\pi_\ell(\theta^*) = \left(\frac{\exp(A_{\ell *}^\top \theta^*)}{1+\exp(A_{\ell *}^\top \theta^*)}\right)$. The resulting optimization problem is as follows:

$$\min_{\theta \in \mathbb{R}^d} \frac{1}{n}\sum_{i=1}^{n}\frac{1}{b}\sum_{\ell \in \mathcal{S}_i}(\log[1 + \exp(A_{\ell *}\theta)] - y_\ell A_{\ell *}\theta) \text{ subject to } \|\theta\|_0 \leq k, \|\theta\|_2 \leq \tau. \tag{3.15}$$

**Remark 3.13.** Due to the additional $\ell_2$-constraint, we need a projection step in SVRG-HT. In particular, we replace Step (S3) in Algorithm 1 with the following update:

$$\theta^{(t+1)} = \Pi_\tau(\mathcal{H}_k(\overline{\theta}^{(t+1)})),$$

where $\Pi_\tau(\cdot) : \mathbb{R}^d \to \mathbb{R}^d$ is an $\ell_2$-norm projection operator defined as $\Pi_\tau(v) = \max\{\|v\|_2, \tau\} \cdot v/\|v\|_2$ for any $v \in \mathbb{R}^d$. Since $\Pi_\tau(\cdot)$ is strictly contractive, i.e., $\|\Pi_\tau(\theta) - \theta^*\|_2 \leq \|\theta - \theta^*\|_2$, Theorem 3.7 still holds[1] for SVRG-HT with this additional projection step.

Assume that for any $v \in \mathbb{R}^d$ with $\|v\|_0 \leq s$ and $\|v\|_2 \leq 2\tau$, the design matrix $A$ satisfies $\frac{\max_j \|A_{*j}\|_2}{\sqrt{nb}} \leq 1$, and the objective $\mathcal{F}(\theta)$ and $\{f_i(\theta)\}_{i=1}^n$ satisfy the RE conditions:

$$v^\top \nabla^2 \mathcal{F}(\theta) v \geq \psi_1 \|v\|_2^2 - \varphi_1 \frac{\log d}{nb} \|v\|_1^2 \quad \text{and} \quad v^\top \nabla^2 f_i(\theta) v \leq \psi_2 \|v\|_2^2 + \varphi_2 \frac{\log d}{b} \|v\|_1^2, \tag{3.16}$$

where $\psi_1$, $\psi_2$, $\varphi_1$ and $\varphi_2$ are constants that do not scale with $(n, b, k^*, d)$ – (3.16) is satisfied by many common examples of sub-Gaussian random design (Loh and Wainwright, 2013). We show that (3.16) implies the RSC and RSS conditions over an $\ell_2$ ball centered at $\theta^*$ with radius $2\tau$.

**Lemma 3.14.** Suppose that $\mathcal{F}(\theta)$ and $\{f_i(\theta)\}_{i=1}^n$ satisfy the RE conditions (3.16). Then, given large enough $n$ and $b$, for any $\theta$ with $\|\theta - \theta^*\|_2 \leq 2\tau$, there exist a constant $C_6$ and an integer $k$ such that $\mathcal{F}(\theta)$ and $\{f_i(\theta)\}_{i=1}^n$ satisfy the RSC and RSS conditions respectively with $s = 2k + k^*$, where

$$k = C_6 k^* \geq C_1 \kappa_s^2 k^*, \quad \rho_s^- \geq \psi_1/2, \quad \text{and} \quad \rho_s^+ \leq 2\psi_2.$$

The proof of Lemma 3.14 is analogous to the proof of Lemma 3.9, thus is omitted. Lemma 3.14 guarantees that the RSC and RSS conditions hold over a neighborhood of $\theta^*$. For sparse GLM, we further assume $\|\theta^*\|_2 \leq \tau$. This implies that for any $\theta \in \mathbb{R}^d$ with $\|\theta\|_2 \leq \tau$, we have $\|\theta - \theta^*\|_2 \leq \|\theta\|_2 + \|\theta^*\|_2 \leq 2\tau$.

Our next result gives the statistical rate of convergence of the obtained estimator for sparse GLM estimation.

**Corollary 3.15.** Suppose that $A_{i*}$'s have i.i.d. sub-Gaussian rows, and $k$, $\eta$ and $m$ are as specified in Theorem 3.7. In addition, suppose $\|\theta^*\|_2 \leq \tau$. Then, given a constant $\delta \in (0,1)$, a sufficiently small accuracy parameter $\varepsilon > 0$, and large enough $n$ and $b$, we need at most $r$ outer iterations given in (3.9) via SVRG-HT to guarantee that, with high probability, we have

$$\|\widetilde{\theta}^{(r)} - \theta^*\|_2 = \mathcal{O}\left(\sqrt{\frac{k^* \log d}{nb}}\right). \tag{3.17}$$

We note that the statistical rate of convergence above matches the state-of-the-art result in parameter estimation for GLM; see Loh and Wainwright (2013) for more details. A proof of Corollary 3.15 is given in Section 7.3.

---
[1] The gap of the objective value is also contractive after projection due to the restricted convexity of $\mathcal{F}$.

### 3.2.3 Low-rank Matrix Recovery

Next, we consider a low-rank matrix linear model

$$y = \mathcal{A}(\Theta^*) + z,$$

where $y \in \mathbb{R}^{nb}$ is the response vector, $\Theta^* \in \mathbb{R}^{d \times p}$ is the unknown low-rank matrix with rank$(\Theta^*) = k^*$, $\mathcal{A}(\cdot) : \mathbb{R}^{d \times p} \to \mathbb{R}^{nb}$ is a linear operator defined as $\mathcal{A}(\Theta) = [\langle A_1, \Theta \rangle, \ldots, \langle A_{nb}, \Theta \rangle]^\top$ for any matrix $\Theta \in \mathbb{R}^{d \times p}$, $A_i \in \mathbb{R}^{d \times p}$ is a measurement matrix for all $i = 1, \ldots, nb$, and $z \in \mathbb{R}^{nb}$ is a random noise vector sampled from $\mathcal{N}(0, \sigma^2 I)$.

As before, we divide the observations into $n$ blocks, indexed by $y_{\mathcal{S}_i}$, where $\mathcal{S}_i$ denotes the corresponding indices of $y$, with $|\mathcal{S}_i| = b$, for all $i = 1, \ldots, n$. Then, the resulting optimization problem is

$$\min_{\Theta \in \mathbb{R}^{d \times p}} \mathcal{F}(\Theta) = \frac{1}{n} \sum_{i=1}^{n} f_i(\Theta) \quad \text{subject to } \text{rank}(\Theta) \leq k, \tag{3.18}$$

where $f_i(\Theta) = \frac{1}{2b} \|y_{\mathcal{S}_i} - \mathcal{A}_{\mathcal{S}_i}(\Theta)\|_2^2$ and $\mathcal{A}_{\mathcal{S}_i}(\Theta)$ denotes a sub-vector of $\mathcal{A}(\Theta)$ indexed by $\mathcal{S}_i$, for all $i = 1, \ldots, n$.

For low-rank matrix problems, we consider the following matrix RSC and RSS conditions that are simple generalization of the RSC and RSS conditions for sparse vectors in Definitions 3.1 and 3.2. These matrix RSC and RSS conditions were studied recently in high-dimensional statistical analyses for low-rank matrix recovery (Negahban and Wainwright, 2011; Negahban et al., 2012; Negahban and Wainwright, 2012).

**Definition 3.16** (Matrix Restricted Strong Convexity Condition). *A differentiable function $\mathcal{F} : \mathbb{R}^{d \times p} \to \mathbb{R}$ is restricted $\rho_s^-$-strongly convex at rank level $s$ if there exists a generic constant $\rho_s^- > 0$ such that for any $\Theta, \Theta' \in \mathbb{R}^{d \times p}$ with $\text{rank}(\Theta - \Theta') \leq s$, we have*

$$\mathcal{F}(\Theta) - \mathcal{F}(\Theta') - \langle \nabla \mathcal{F}(\Theta'), \Theta - \Theta' \rangle \geq \frac{\rho_s^-}{2} \|\Theta - \Theta'\|_F^2. \tag{3.19}$$

**Definition 3.17** (Matrix Restricted Strong Smoothness Condition). *For any $i \in [n]$, a differentiable function $f_i : \mathbb{R}^{d \times p} \to \mathbb{R}$ is restricted $\rho_s^+$-strongly smooth at rank level $s$ if there exists a generic constant $\rho_s^+ > 0$ such that for any $\Theta, \Theta' \in \mathbb{R}^{d \times p}$ with $\text{rank}(\Theta - \Theta') \leq s$, we have*

$$f_i(\Theta) - f_i(\Theta') - \langle \nabla f_i(\Theta'), \Theta - \Theta' \rangle \leq \frac{\rho_s^+}{2} \|\Theta - \Theta'\|_F^2. \tag{3.20}$$

As with the RSC and RSS conditions, the matrix RSC and RSS conditions can be verified for $\mathcal{F}(\Theta)$ and $\{f_i(\Theta)\}_{i=1}^{n}$ by studying sub-Gaussian random design (Negahban and Wainwright, 2011). Specifically, if $\{A_i\}_{i=1}^{nb}$ in the linear operator $\mathcal{A}(\cdot)$ are drawn i.i.d. from the $\Sigma_\mathcal{A}$-Gaussian ensemble, i.e., $\text{vec}(A_i) \sim \mathcal{N}(0, \Sigma_\mathcal{A})$ with $\Sigma_\mathcal{A} \in \mathbb{R}^{dp \times dp}$, then, with high probability, we have

$$\frac{\mathcal{A}(\Theta)}{\sqrt{nb}} \geq \psi_1 \|\sqrt{\Sigma_\mathcal{A}} \text{vec}(\Theta)\|_2 - \varphi_1 \rho(\Sigma_\mathcal{A}) \left( \sqrt{\frac{d}{nb}} + \sqrt{\frac{p}{nb}} \right) \|\Theta\|_* \quad \text{and}$$

$$\frac{\mathcal{A}_{\mathcal{S}_i}(\Theta)}{\sqrt{b}} \leq \psi_2 \|\sqrt{\Sigma_\mathcal{A}} \text{vec}(\Theta)\|_2 - \varphi_2 \rho(\Sigma_\mathcal{A}) \left( \sqrt{\frac{d}{b}} + \sqrt{\frac{p}{b}} \right) \|\Theta\|_* \quad \text{for all } i = 1, \ldots, n,$$

where $\rho^2(\Sigma_{\mathcal{A}}) = \sup_{\|u\|_2=1, \|v\|_2=1} \text{var}(u^\top X v)$, and the random matrix $X$ is sampled from the $\Sigma_{\mathcal{A}}$-Gaussian ensemble. This further implies that $\mathcal{F}(\Theta)$ and $\{f_i(\Theta)\}_{i=1}^n$ satisfy the matrix RSC and RSS conditions respectively for large enough $k$, following the result in Lemma 3.14.

**Remark 3.18** (SVRG-HT for Singular Value Thresholding). For low-rank matrix recovery, we need to replace the hard thresholding operator $\mathcal{H}_k(\cdot)$ in Step (S3) of Algorithm 1 by the singular value thresholding operator $\mathcal{R}_k(\cdot)$. In particular, we replace Step (S3) with the following update:

$$\Theta^{(t+1)} = \mathcal{R}_k(\overline{\Theta}^{(t+1)}) = \sum_{i=1}^{k} \overline{\sigma}_i \overline{U}_i \overline{V}_i^\top,$$

where $\overline{\sigma}_i$, $\overline{U}_i$, and $\overline{V}_i$ are the $i$-th largest singular value, and the corresponding left and right singular vectors of $\overline{\Theta}^{(t+1)}$ respectively.

For sparse vectors, Lemma 3.3 guarantees that the hard thresholding operation is nearly nonexpansive when $k$ is sufficiently larger than $k^*$. We provide a similar result for the singular value thresholding operation on matrices.

**Lemma 3.19.** Recall that $\Theta^* \in \mathbb{R}^{d \times p}$ is the unknown low-rank matrix of interest with $\text{rank}(\Theta^*) \leq k^*$, and $\mathcal{R}_k(\cdot) : \mathbb{R}^{d \times p} \to \mathbb{R}^{d \times p}$ is the singular value thresholding operator, which keeps the largest $k$ singular values and sets the other singular values equal to zero. Given $k > k^*$, for any matrix $\Theta \in \mathbb{R}^{d \times p}$, we have

$$\|\mathcal{R}_k(\Theta) - \Theta^*\|_F^2 \leq \left(1 + \frac{2\sqrt{k^*}}{\sqrt{k - k^*}}\right) \cdot \|\Theta - \Theta^*\|_F^2. \quad (3.21)$$

See Appendix 9.5 for a proof of Lemma 3.19. Given Lemma 3.19, the computational theory follows directly from Theorem 3.7. This further allows us to characterize the statistical properties of the obtained estimator for low-rank matrix recovery as follows.

**Corollary 3.20.** Suppose that in the linear operator $\mathcal{A}(\cdot)$, $\text{vec}(A_i)$ is drawn i.i.d. from $\mathcal{N}(0, \Sigma_{\mathcal{A}})$, and $k$, $\eta$ and $m$ are as specified in Theorem 3.7. Then, given a constant $\delta \in (0, 1)$, a sufficiently small accuracy parameter $\varepsilon > 0$, and large enough $n$ and $b$, we need at most $r$ outer iterations given in (3.9) via SVRG-HT to guarantee that, with high probability, we have

$$\|\widetilde{\Theta}^{(r)} - \Theta^*\|_F = \mathcal{O}\left(\sigma \sqrt{\frac{k^*(d+p)}{nb}}\right).$$

The statistical rate of the convergence in Corollary 3.20 matches with the state-of-the-art result in parameter estimation for low-rank matrix recovery (Negahban and Wainwright, 2011). The analysis follows directly from Corollary 3.10 and Negahban and Wainwright (2011).

# 4 Asynchronous SVRG-HT

We extend SVRG-HT to an asynchronous parallel variant, named asynchronous SVRG-HT (ASVRG-HT). Here, we assume a parallel computing procedure with a multicore architecture, where each

---

**Algorithm 3** Asynchronous Stochastic Variance Reduced Gradient Hard Thresholding Algorithm. We assume a parallel computing procedure with a multicore architecture, where each processor makes a stochastic gradient update of a global parameter stored in a shared memory via an asynchronous and lock-free mode.

---

**Input:** update frequency $m$, step size parameter $\eta$, sparsity $k$, and initial solution $\widetilde{\theta}^{(0)}$
for $r = 1, 2, \ldots$ do
    $\widetilde{\theta} = \widetilde{\theta}^{(r-1)}$
    $\widetilde{\mu} = \frac{1}{n}\sum_{i=1}^{n}\nabla f_i(\widetilde{\theta})$
    $\theta^{(0)} = \widetilde{\theta}$
    for $t = 0, 1, \ldots, m-1$ do
        (S1) Randomly sample $i_t$ from $[n]$ and $e_t \subset [d]$ with $|e_t| \leq k$
        (S2) $\overline{\theta}^{(t+1)} = \theta^{(t)} - \eta \cdot [g^{(t)}(\theta^{(t)})]_{e_t}$, where $g^{(t)}(\theta^{(t)}) = \nabla f_{i_t}(\theta^{(t)}) - \nabla f_{i_t}(\widetilde{\theta}) + \widetilde{\mu}$
        (S3) $\theta^{(t+1)} = \mathcal{H}_k(\overline{\theta}^{(t+1)})$
    end for
    $\widetilde{\theta}^{(r)} = \theta^{(t+1)}$ for randomly chosen $t \in [m]$
end for

---

processor makes a stochastic gradient update on a global parameter stored in a shared memory in an asynchronous and lock-free mode. This setup is similar to that considered in many asynchronous algorithms (Recht et al., 2011; Reddi et al., 2015; Liu et al., 2015; Mania et al., 2015).

The asynchronous version differs from the original as follows: at the $t$-th iteration of inner loop of ASVRG-HT, we sample an index $i_t \in [n]$ of the component function uniformly randomly, and sample an index set $e_t \subset [d]$ over all subsets of $[d]$ of size bounded by $k$, also uniformly randomly. The parameter vector $\theta^{(t)}$ is updated only over the sampled index set $e_t$; see Algorithm 3 for more details.

In order to formally analyze ASVRG-HT, we introduce two parameters capturing the key notions of parallelism and data sparsity in asynchronous updates (Recht et al., 2011). The first parameter, $\varsigma$, captures the degree of parallelism in the asynchronous algorithm. Let $t'$ be the actual $\theta$-iterate when evaluation is performed at the $t$-th iteration, then $\varsigma$ is the smallest positive integer such that $t - t' \leq \varsigma$ for any $t$. This upper bound on the delay characterizes the degree of parallelism in the asynchronous method. The more parallel computations are adopted, the larger value of $\varsigma$ can be. The value of $\varsigma$ is approximately linear on the number of cores in parallel computing architecture (Recht et al., 2011; Liu et al., 2015).

The second parameter, $\Delta$, captures the sparsity of data. Suppose $f_i(\theta)$ only depends on $\theta_{e_i}$, where $e_i \subset [d]$ and $|e_i| = k_i$ for some positive integer $k_i$. Then, $\Delta \in [0, 1]$ is the smallest constant such that $\mathbb{E}\|\theta_e\|_2^2 \leq \Delta\|\theta\|_2^2$, where $e \subseteq [d]$ is a subset of $[d]$ sampled uniformly randomly from sets with cardinality $|e| = k_i$. The sparser $\theta$ is, on which $f_i$ depends, the smaller $\Delta$ is. We are interested in the setting $\Delta \ll 1$.

The following Theorem characterizes the error of the objective value and estimation error for ASVRG-HT.

**Theorem 4.1.** Let $\theta^*$ denote the unknown sparse parameter vector of the underlying statistical model, with $\|\theta^*\|_0 \leq k^*$. Assume that the objective function $\mathcal{F}(\theta)$ satisfies the RSC condition and functions $\{f_i(\theta)\}_{i=1}^n$ satisfy the RSS condition with $s = 2k + k^*$, where $k \geq C_1 \kappa_s^2 k^*$ and $C_1$ is a generic constant. Define

$$\widetilde{\mathcal{I}} = \mathrm{supp}(\mathcal{H}_{2k}(\nabla \mathcal{F}(\theta^*))) \cup \mathrm{supp}(\theta^*).$$

Then, there exist generic constants $C_2, C_3, C_4$, and $C_5$ such that by setting $\eta \rho_s^+ \in [C_2, C_3]$, $m \geq C_4 \kappa_s$ and $\Delta \varsigma^2 \leq C_5$, we get

$$\frac{\left(1 + \frac{2\sqrt{k^*}}{\sqrt{k-k^*}}\right)^m \cdot \frac{2\sqrt{k^*}}{\sqrt{k-k^*}}}{\eta \rho_s^- (1 - 12 \eta \rho_s^+ \Gamma)\left(\left(1 + \frac{2\sqrt{k^*}}{\sqrt{k-k^*}}\right)^m - 1\right)} + \frac{12 \eta \rho_s^+ \Gamma}{1 - 12 \eta \rho_s^+ \Gamma} \leq \frac{5}{6},$$

where $\Gamma = \frac{1 + \rho_s^+ \Delta \varsigma^2 \eta}{1 - 2\rho_s^{+2} \Delta \varsigma^2 \eta^2}$. Furthermore, the parameter $\widetilde{\theta}^{(r)}$ at the $r$-th iteration of ASVRG-HT satisfies

$$\mathbb{E}\left[\mathcal{F}(\widetilde{\theta}^{(r)}) - \mathcal{F}(\theta^*)\right] \leq \left(\frac{5}{6}\right)^r \left[\mathcal{F}(\widetilde{\theta}^{(0)}) - \mathcal{F}(\theta^*)\right] + \frac{18 \eta \Gamma}{1 - 12 \eta \rho_s^+ \Gamma} \|\nabla_{\widetilde{\mathcal{I}}} \mathcal{F}(\theta^*)\|_2^2$$
$$+ 2\sqrt{s} \|\nabla \mathcal{F}(\theta^*)\|_\infty \mathbb{E}\|\widetilde{\theta}^{(r-1)} - \theta^*\|_2 \text{ and}$$

$$\mathbb{E}\|\widetilde{\theta}^{(r)} - \theta^*\|_2 \leq \sqrt{\frac{2\left(\frac{5}{6}\right)^r \left[\mathcal{F}(\widetilde{\theta}^{(0)}) - \mathcal{F}(\theta^*)\right]}{\rho_s^-}} + \frac{\sqrt{s} \|\nabla \mathcal{F}(\theta^*)\|_\infty}{\rho_s^-} + \|\nabla_{\widetilde{\mathcal{I}}} \mathcal{F}(\theta^*)\|_2 \sqrt{\frac{36 \eta \Gamma}{(1 - 12 \eta \rho_s^+ \Gamma) \rho_s^-}}$$
$$+ \sqrt{2 \frac{\sqrt{s} \|\nabla \mathcal{F}(\theta^*)\|_\infty \mathbb{E}\|\widetilde{\theta}^{(r-1)} - \theta^*\|_2}{\rho_s^-}}.$$

Moreover, given a constant $\delta \in (0, 1)$ and a pre-specified accuracy $0 < \varepsilon < \frac{2\sqrt{s} \|\nabla \mathcal{F}(\theta^*)\|_\infty}{\rho_s^-}$, we need at most

$$r = \left\lceil 4 \log\left(\frac{\mathcal{F}(\widetilde{\theta}^{(0)}) - \mathcal{F}(\theta^*)}{\varepsilon \delta}\right) + 2 \log \frac{\left|\|\widetilde{\theta}^{(0)} - \theta^*\|_2 - \frac{7\sqrt{s} \|\nabla \mathcal{F}(\theta^*)\|_\infty}{\rho_s^-}\right|}{\varepsilon \delta} \right\rceil$$

outer iterations to guarantee that with probability at least $1 - 2\delta$, we have simultaneously

$$\mathcal{F}(\widetilde{\theta}^{(r)}) - \mathcal{F}(\theta^*) \leq \varepsilon \left(1 + 2\sqrt{s} \|\nabla \mathcal{F}(\theta^*)\|_\infty\right) + \frac{18 \eta \Gamma}{1 - 12 \eta \rho_s^+ \Gamma} \|\nabla_{\widetilde{\mathcal{I}}} \mathcal{F}(\theta^*)\|_2^2 \text{ and}$$

$$\|\widetilde{\theta}^{(r)} - \theta^*\|_2 \leq \sqrt{\frac{2\varepsilon\left(1 + 2\sqrt{s} \|\nabla \mathcal{F}(\theta^*)\|_\infty\right)}{\rho_s^-}} + \frac{\sqrt{s} \|\nabla \mathcal{F}(\theta^*)\|_\infty}{\rho_s^-} + \|\nabla_{\widetilde{\mathcal{I}}} \mathcal{F}(\theta^*)\|_2 \sqrt{\frac{36 \eta \Gamma}{(1 - 12 \eta \rho_s^+ \Gamma) \rho_s^-}}.$$

Theorem 4.1 indicates that ASVRG-HT has a similar iteration complexity to SVRG-HT. Therefore, when $\Delta \varsigma^2 = \mathcal{O}(1)$, ASVRG-HT can be $\varsigma$ times faster than SVRG-HT due to the parallelism. For example, if $\Delta = \mathcal{O}(k/d)$, then we achieve a speedup of $\varsigma = \Omega(\sqrt{d/k})$ times, which is analogous to ASVRG in Reddi et al. (2015). A proof of Theorem 4.1 is presented in Section 7.4.

From the computational guarantees for ASVRG-HT in Theorem 4.1, we can further establish statistical guarantees for popular constrained M-estimation problems. Given proper conditions on the design, as in Section 3.2, we can demonstrate the optimal statistical rate of convergence for sparse linear regression, sparse generalized linear model estimation, and low-rank matrix estimation. The analysis follows directly from the statistical theory of SVRG-HT in Section 3.2.

**Remark 4.2.** For the sake of convenience, in our analysis we assumed sampling with-replacement of indices as well as the index set. However, in practice, sampling without-replacement can perhaps significantly improve the efficiency, as also noted in previous work (Recht et al., 2011).

# 5  Experiments

We compare the empirical performance of SVRG-HT with two other competitors: FG-HT proposed in Jain et al. (2014) and SG-HT proposed in Nguyen et al. (2014) on both synthetic data and real data. We also compare the performance of parameter estimation between the $\ell_0$-constrained problem (1.1) and an $\ell_1$-regularized problem solved by the proximal stochastic variance reduced gradient (Prox-SVRG) algorithm (Xiao and Zhang, 2014).

## 5.1  Synthetic Data

We consider a sparse linear regression problem. We generate each row of the design matrix $A_{i*}$, $i \in [nb]$, independently from a $d$-dimensional Gaussian distribution with mean 0 and covariance matrix $\Sigma \in \mathbb{R}^{d \times d}$. The response vector is generated from the linear model $y = A\theta^* + z \in \mathbb{R}^{nb}$, where $\theta^* \in \mathbb{R}^d$ is the $k^*$-sparse regression coefficient vector, and $z$ is generated from an $n$-dimensional Gaussian distribution with mean 0 and covariance matrix $\sigma^2 I$. We set $nb = 10000$, $d = 25000$, $k^* = 200$ and $k = 500$. For $\Sigma$, we set $\Sigma_{ii} = 1$ and $\Sigma_{ij} = c$ for some constant $c \in (0,1)$ for all $i \neq j$. The nonzero entries in $\theta^*$ are sampled independently from a uniform distribution over the interval $(-2, +2)$. We divide 10000 samples into $n$ mini batches, and each mini batch contains $b = 10000/n$ samples.

Figure 1 illustrates the computational performance of FG-HT, SG-HT, and SVRG-HT for eight different settings of $(n, b)$ and $\Sigma_{ij}$, each with step sizes $\eta = 1/256, 1/512$, and $1/1024$. The first four settings are noiseless, i.e., $\sigma = 0$ with (1) $(n, b) = (10000, 1)$, $\Sigma_{ij} = 0.1$; (2) $(n, b) = (10000, 1)$, $\Sigma_{ij} = 0.5$; (3) $(n, b) = (200, 50)$, $\Sigma_{ij} = 0.1$; (4) $(n, b) = (200, 50)$, $\Sigma_{ij} = 0.5$. For simplicity, we choose the update frequency of the inner loop as $m = n$ throughout our experiments[2]. The last four settings are noisy with $\sigma = 1$ and identical choices of $(n, b)$, $\Sigma_{ij}$ and $m$ as in (1)-(4). For all algorithms, we plot the objective values averaged over 50 different runs. The horizontal axis corresponds to the number of passes over the entire dataset; computing a full gradient is counted as 1 pass, while computing a stochastic gradient is counted as $1/n$-th of a pass. The vertical axis corresponds to the ratio of current objective value over the objective value using $\widetilde{\theta}^{(0)} = 0$. We further provide the optimal relative estimation error $\|\widetilde{\theta}^{(10^6)} - \theta^*\|_2 / \|\theta^*\|_2$ after $10^6$ effective passes of the entire dataset for all settings of the three algorithms in Table 1. The estimation error is obtained by averaging over 50 different runs, each of which is chosen from a sequence of step sizes $\eta \in \{1/2^5, 1/2^6, \ldots, 1/2^{14}\}$.

---
[2]Larger $m$ results in increasing number of effective passes of the entire dataset required to achieve the same decrease of objective values, which is also observed in Prox-SVRG (Xiao and Zhang, 2014)

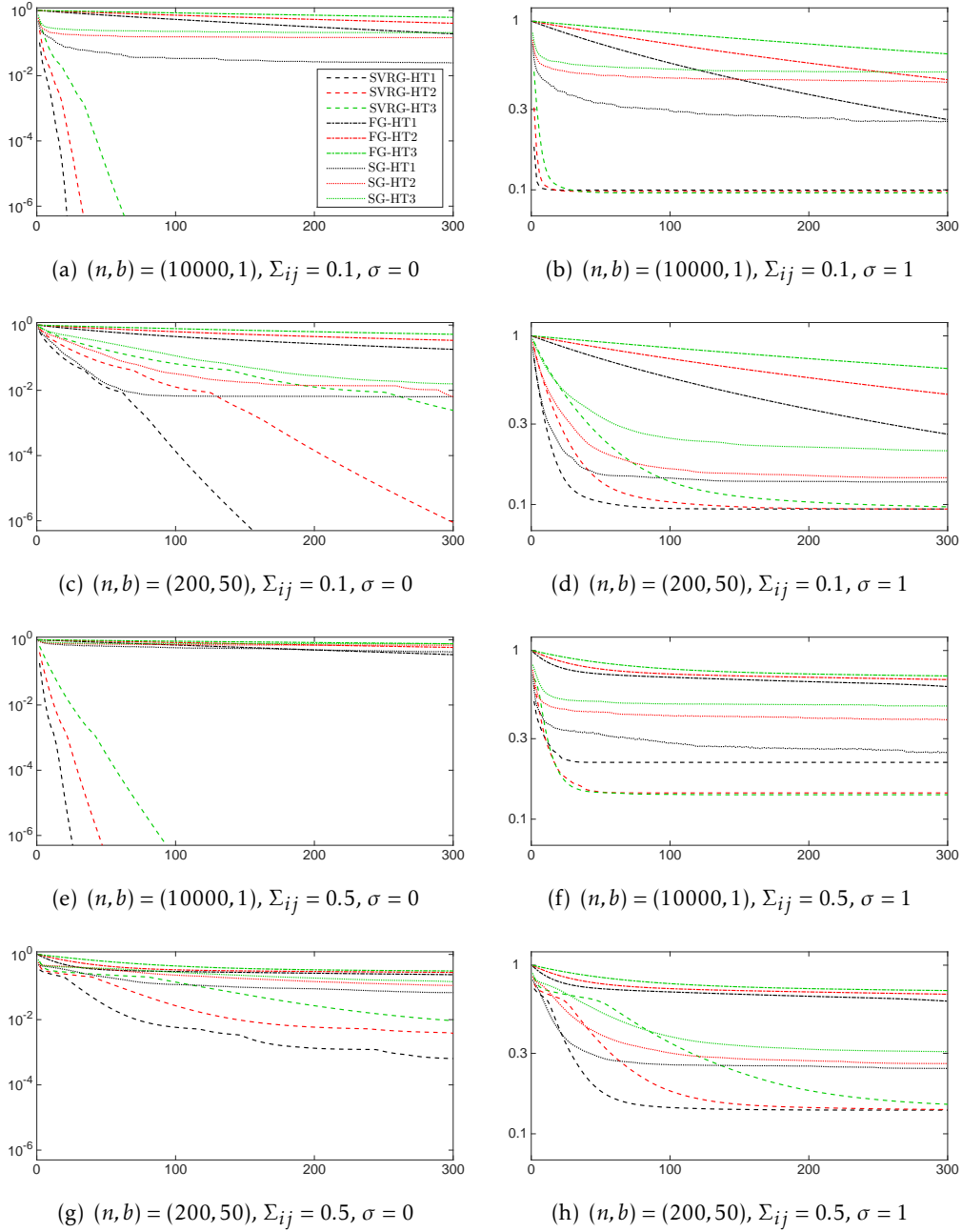

Figure 1: Comparison among the three algorithms in all settings on the simulated data. The horizontal axis corresponds to the number of passes over the entire dataset. The vertical axis corresponds to the ratio of current objective value over the objective value using $\widetilde{\theta}^{(0)} = 0$. For each algorithm, option 1, 2 and 3 correspond to the step sizes $\eta = 1/256, 1/512$, and $1/1024$ respectively. It is evident from the plots that SVRG-HT outperforms the other competitors in terms of the convergence rate over all settings.

Table 1: Comparison of optimal relative estimation errors among the three algorithms in all settings on the simulated data. We denote $(n,b)_1 = (10000,1)$ and $(n,b)_2 = (200,50)$. SVRG-HT achieves comparable result with FG-HT, both of which outperforms SG-HT over all settings.

| Method | $\sigma = 0$ | | | | $\sigma = 1$ | | | |
| --- | --- | --- | --- | --- | --- | --- | --- | --- |
| | $\Sigma_{ij} = 0.1$ | | $\Sigma_{ij} = 0.5$ | | $\Sigma_{ij} = 0.1$ | | $\Sigma_{ij} = 0.5$ | |
| | $(n,b)_1$ | $(n,b)_2$ | $(n,b)_1$ | $(n,b)_2$ | $(n,b)_1$ | $(n,b)_2$ | $(n,b)_1$ | $(n,b)_2$ |
| FG-HT | $< 10^{-20}$ | | $< 10^{-20}$ | | 0.00851 | | 0.02940 | |
| SG-HT | $< 10^{-20}$ | $< 10^{-20}$ | $< 10^{-20}$ | 0.13885 | 0.02490 | 0.06412 | 0.21676 | 0.18764 |
| SVRG-HT | $< 10^{-20}$ | $< 10^{-20}$ | $< 10^{-20}$ | $< 10^{-20}$ | 0.00968 | 0.00970 | 0.02614 | 0.02823 |

We see from Figure 1 that SVRG-HT outperforms the other competitors in terms of the convergence rate in all settings. While FG-HT also enjoys linear converge guarantees, its computational cost at each iteration is $n$ times larger than that of SVRG-HT. Consequently, its performance is much worse than that of SVRG-HT. Besides, we also see that SG-HT converges slower than SVRG-HT in all settings. This is because the largest eigenvalue of any 500 by 500 submatrix of the covariance matrix is large (larger than 50 or 250) such that the underlying design matrix violates the Restricted Isometry Property (RIP) required by SG-HT. On the other hand, Table 1 indicates that the optimal estimation error of SVRG-HT is comparable to FG-HT, both of which outperform SG-HT, especially in noisy settings. It is important to note that with the optimal step size, the estimation of FG-HT usually becomes stable after $> 10^5$ passes, while the estimation of SVRG-HT usually becomes stable within a few dozen to a few hundred passes, which validates the significant improvement of SVRG-HT over FG-HT in terms of the computational cost.

## 5.2 Real Data

We adopt a subset of RCV1 dataset with 9625 documents and 29992 distinct words, including the classes of "C15", "ECAT", "GCAT", and "MCAT" (Cai and He, 2012). We apply logistic regression to perform a binary classification for all classes, each of which uses 5000 documents for training, i.e., $nb = 5000$ and $d = 29992$, with the same proportion of documents from each class, and the rest for testing. We illustrate the computational performance of FG-HT, SG-HT, and SVRG-HT in two different settings: Setting (1) has $(n,b) = (5000,1)$; Setting (2) has $(n,b) = (100,50)$. We choose $k = 200$ and $m = n$ for both settings. For all three algorithms, we plot their objective values and provide the optimal classification errors averaged over 10 different runs using random data separations. Figure 2 demonstrates the computational performance for "C15" on the training dataset, and the other classes have similar performance. The horizontal axis corresponds to the number of passes over the entire training dataset. The vertical axis corresponds to the ratio of current objective value over the initial objective value using $\widetilde{\theta}^{(0)} = 0$. Similar to the synthetic data, SVRG-HT outperforms the other competitors in terms of the convergence rate in both settings.

We further provide the optimal misclassification rates of all classes for the three algorithms in Table 2, where the optimal step size $\eta$ for each algorithm is chosen from a sequence of values $\{1/2^5, 1/2^6, \ldots, 1/2^{14}\}$. Similar to the synthetic data again, the optimal misclassification rate of SVRG-HT is comparable to FG-HT, both of which outperform SG-HT. The estimation of FG-HT generally requires $> 10^6$ passes to become stable, while the estimation of SVRG-HT generally requires a few hundred to a few thousand passes to be stable, which validates the significant improvement of SVRG-HT over FG-HT on this real dataset in terms of the computational cost.

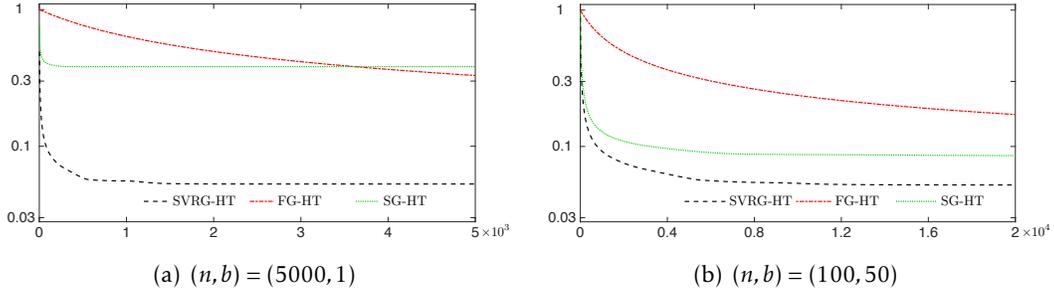

(a) $(n,b) = (5000,1)$

(b) $(n,b) = (100,50)$

Figure 2: Comparison among the three algorithms in two different settings on the training dataset of RCV1 for the class "C15". The horizontal axis corresponds to the number of passes over the entire training dataset. The vertical axis corresponds to the ratio of current objective value over the initial objective. It is evident from the plots that SVRG-HT outperforms the other competitors in both settings.

Table 2: Comparison of optimal classification errors on the test dataset of RCV1 among the three algorithms for both settings and all four classes. We denote $(n,b)_1 = (5000,1)$ and $(n,b)_2 = (100,50)$. SVRG-HT achieves comparable result with FG-HT, both of which outperform SG-HT over all settings.

|  | C15 | | ECAT | | GCAT | | MCAT | |
|---|---|---|---|---|---|---|---|---|
|  | $(n,b)_1$ | $(n,b)_2$ | $(n,b)_1$ | $(n,b)_2$ | $(n,b)_1$ | $(n,b)_2$ | $(n,b)_1$ | $(n,b)_2$ |
| FG-HT | 0.02844 | | 0.05581 | | 0.03028 | | 0.05703 | |
| SG-HT | 0.03259 | 0.03361 | 0.06851 | 0.07179 | 0.06263 | 0.09142 | 0.07638 | 0.08228 |
| SVRG-HT | 0.02826 | 0.02867 | 0.05628 | 0.05631 | 0.03354 | 0.03444 | 0.05877 | 0.05927 |

### 5.3 $\ell_0$-Norm/SVRG-HT vs. $\ell_1$-Norm/Prox-SVRG

We further discuss the empirical performance of sparsity induced problems using the $\ell_0$-norm and the $\ell_1$-norm respectively. Specifically, we consider the sparse linear regression problem (2.2) for the $\ell_0$-constrained problem and the following $\ell_1$-regularized problem,

$$\min_{\theta \in \mathbb{R}^d} \frac{1}{n} \sum_{i=1}^{n} \frac{1}{b} \|y_{\mathcal{S}_i} - A_{\mathcal{S}_i *}\theta\|_2^2 + \lambda \|\theta\|_1, \tag{5.1}$$

where $\lambda > 0$ is a regularization parameter. The $\ell_0$-constrained problem (2.2) is solved by SVRG-HT, and the $\ell_1$-regularized problem (5.1) is solved by Prox-SVRG (Xiao and Zhang, 2014). We follow the same settings as in Section 5.1 for data generation and the choice of parameters for SVRG-HT. For the $\ell_1$-regularized problem (5.1), we choose an optimal regularization parameter $\lambda$ from a sequence of values $\{1/2^2, 1/2^4, 1/2^6, \ldots, 1/2^{20}\}$, which returns the optimal relative estimation error $\|\widetilde{\theta}^{(10^6)} - \theta^*\|_2 / \|\theta^*\|_2$.

Table 3: Comparison of optimal relative estimation errors between (2.2) and (5.1) in all settings on the synthetic data. We denote $(n,b)_1 = (10000, 1)$ and $(n,b)_2 = (200, 50)$.

| Method | $\sigma = 0$ | | | | $\sigma = 1$ | | | |
| --- | --- | --- | --- | --- | --- | --- | --- | --- |
| | $\Sigma_{ij} = 0.1$ | | $\Sigma_{ij} = 0.5$ | | $\Sigma_{ij} = 0.1$ | | $\Sigma_{ij} = 0.5$ | |
| | $(n,b)_1$ | $(n,b)_2$ | $(n,b)_1$ | $(n,b)_2$ | $(n,b)_1$ | $(n,b)_2$ | $(n,b)_1$ | $(n,b)_2$ |
| $\ell_0$-norm | $< 10^{-20}$ | $< 10^{-20}$ | $< 10^{-20}$ | $< 10^{-20}$ | 0.00968 | 0.00970 | 0.02614 | 0.02823 |
| $\ell_1$-norm | $\approx 10^{-6}$ | $\approx 10^{-7}$ | $\approx 10^{-6}$ | $\approx 10^{-7}$ | 0.01715 | 0.01306 | 0.08475 | 0.08177 |

Table 3 provides the optimal estimation errors in all settings, each of which is averaged over 50 different runs. We observe that the $\ell_0$-norm problem uniformly outperforms the $\ell_1$-norm problem in terms of statistical accuracy. Besides, it is important to note that we only need to tune the step size $\eta$ for the $\ell_0$-norm problem (2.2), which is insensitive in different settings, and the sparsity parameter $k$ is fixed throughout. On the other hand, for the $\ell_1$-norm problem (5.1), we need to tune both the step size $\eta$ and the regularization parameter $\lambda$ to obtain the optimal estimation, which require much more tuning efforts. Moreover, we observe that SVRG-HT converges faster than Prox-SVRG, where SVRG-HT typically requires a few dozen to a few hundred passes of data to converge. This is because SVRG-HT always guarantees the solution sparsity, and the restricted strong convexity enables the fast convergence. In contrast, Prox-SVRG requires a few thousand passes of data to converge, because Prox-SVRG often yields dense solutions, especially at the first few iterations.

## 6 Discussion

We provide a summary of comparison between our proposed algorithm SVRG-HT with FG-HT (Jain et al., 2014) and SG-HT (Nguyen et al., 2014) in Table 4. We want to remark that though the computational complexity of SG-HT may seem lower than SVRG-HT, the RSC and RSS conditions of SG-HT are very restrictive, and it generally converges much slower than SVRG-HT in practice.

SVRG-HT is closely related to some recent work on stochastic optimization algorithms, including Prox-SVRG (Xiao and Zhang, 2014), stochastic averaging gradient (SAG) (Roux et al., 2012) and stochastic dual coordinate ascent (SDCA) (Shalev-Shwartz and Zhang, 2013). However, the focus in these previous works has been on establishing global linear convergence for optimization problems involving strongly convex objective with a convex constraint, whereas SVRG-HT guarantees linear convergence for optimization problems involving a nonconvex objective with nonconvex cardinality constraint.

Table 4: Comparison with FG-HT (Jain et al., 2014) and SG-HT (Nguyen et al., 2014). Our contributions are manifold: (1) less restrictive assumptions on the RSC and RSS conditions than SG-HT; (2) improving the iteration complexity and computational complexity over FG-HT; and (3) improving the statistical performance over SG-HT. We only provide the statistical error of sparse linear regression for illustration.

| Method | Restrictions on $\kappa_s$ | Iteration Complexity | Comput. Complexity | Stat. Error |
| --- | --- | --- | --- | --- |
| FG-HT | No: $\kappa_s$ bounded | $\mathcal{O}(\kappa_s \log(1/\varepsilon))$ | $\mathcal{O}(nb\kappa_s \cdot \log(1/\varepsilon))$ | $\mathcal{O}\left(\sigma\sqrt{k^* \log d/(nb)}\right)$ |
| SG-HT | Yes: $\kappa_s \leq \frac{4}{3}$ | $\mathcal{O}(\log(1/\varepsilon))$ | $\mathcal{O}(\log(1/\varepsilon))$ | $\mathcal{O}\left(\sigma\sqrt{k^* \log d/b}\right)$ |
| SVRG-HT | No: $\kappa_s$ bounded | $\mathcal{O}(\log(1/\varepsilon))$ | $\mathcal{O}([nb + \kappa_s] \cdot \log(1/\varepsilon))$ | $\mathcal{O}\left(\sigma\sqrt{k^* \log d/(nb)}\right)$ |

Other related work includes nonconvex regularized M-estimators proposed in Loh and Wainwright (2013). In particular, the following nonconvex optimization problem is considered in Loh and Wainwright (2013):

$$\min_{\theta} \mathcal{F}(\theta) + \mathcal{P}_{\lambda,\gamma}(\theta) \quad \text{subject to } \|\theta\|_1 \leq R, \tag{6.1}$$

where $\mathcal{P}_{\lambda,\gamma}(\theta)$ is a nonconvex regularization function with tuning parameters $\lambda$ and $\gamma$; Popular choices for $\mathcal{P}_{\lambda,\gamma}(\theta)$ are the SCAD and MCP regularization functions studied in Fan and Li (2001); Zhang (2010). It is shown in Loh and Wainwright (2013) that under restricted strong convexity and restricted strong smoothness conditions, similar to those studied here, the proximal gradient descent attains linear convergence to approximate global optima with optimal estimation accuracy. Accordingly, one could adopt the Prox-SVRG to solve (6.1) in a stochastic fashion, and trim the analyses in Xiao and Zhang (2014) and Loh and Wainwright (2013) to establish similar convergence guarantees. We remark, however, that Problem (6.1) involves three tuning parameters, $\lambda$, $\gamma$, and $R$ which, in practice, requires a large amount of tuning effort to attain good empirical performance. In contrast, Problem (1.1) involves a single tuning parameter, $k$, which makes tuning more efficient.

# 7 Proofs of Main Results

We present the proofs of our main theoretical results in this section.

## 7.1 Proof of Theorem 3.7

**Part 1**. We first demonstrate (3.7) and (3.8). let $v = \theta^{(t)} - \eta g_{\mathcal{I}}^{(t)}(\theta^{(t)})$ and $\mathcal{I} = \mathcal{I}^* \cup \mathcal{I}^{(t)} \cup \mathcal{I}^{(t+1)}$, where $\mathcal{I}^* = \text{supp}(\theta^*)$, $\mathcal{I}^{(t)} = \text{supp}(\theta^{(t)})$ and $\mathcal{I}^{(t+1)} = \text{supp}(\theta^{(t+1)})$. Conditioning on $\theta^{(t)}$, we have the

following expectation:

$$\begin{aligned}
\mathbb{E}\|v - \theta^*\|_2^2 &= \mathbb{E}\|\theta^{(t)} - \eta g_{\mathcal{I}}^{(t)}(\theta^{(t)}) - \theta^*\|_2^2 \\
&= \mathbb{E}\|\theta^{(t)} - \theta^*\|_2^2 + \eta^2 \mathbb{E}\|g_{\mathcal{I}}^{(t)}(\theta^{(t)})\|_2^2 - 2\eta\langle\theta^{(t)} - \theta^*, \mathbb{E} g_{\mathcal{I}}^{(t)}(\theta^{(t)})\rangle \\
&= \mathbb{E}\|\theta^{(t)} - \theta^*\|_2^2 + \eta^2 \mathbb{E}\|g_{\mathcal{I}}^{(t)}(\theta^{(t)})\|_2^2 - 2\eta\langle\theta^{(t)} - \theta^*, \nabla_{\mathcal{I}}\mathcal{F}(\theta^{(t)})\rangle \\
&\leq \mathbb{E}\|\theta^{(t)} - \theta^*\|_2^2 + \eta^2 \mathbb{E}\|g_{\mathcal{I}}^{(t)}(\theta^{(t)})\|_2^2 - 2\eta\left[\mathcal{F}(\theta^{(t)}) - \mathcal{F}(\theta^*)\right] \\
&\leq \mathbb{E}\|\theta^{(t)} - \theta^*\|_2^2 - 2\eta\left[\mathcal{F}(\theta^{(t)}) - \mathcal{F}(\theta^*)\right] \\
&\quad + 12\eta^2 \rho_s^+ \left[\mathcal{F}(\theta^{(t)}) - \mathcal{F}(\theta^*) + \mathcal{F}(\widetilde{\theta}) - \mathcal{F}(\theta^*)\right] + 3\eta^2 \|\nabla_{\mathcal{I}}\mathcal{F}(\theta^*)\|_2^2 \\
&= \mathbb{E}\|\theta^{(t)} - \theta^*\|_2^2 - 2\eta(1 - 6\eta\rho_s^+)\left[\mathcal{F}(\theta^{(t)}) - \mathcal{F}(\theta^*)\right] \\
&\quad + 12\eta^2 \rho_s^+ \left[\mathcal{F}(\widetilde{\theta}) - \mathcal{F}(\theta^*)\right] + 3\eta^2 \|\nabla_{\mathcal{I}}\mathcal{F}(\theta^*)\|_2^2, \quad (7.1)
\end{aligned}$$

where the first inequality follows from the restricted convexity of $\mathcal{F}(\theta)$ and the fact that $\|(\theta^{(t)} - \theta^*)_{\mathcal{I}^c}\|_0 = 0$, and the second inequality follows from Lemma 3.5.

Since $\theta^{(t+1)} = \overline{\theta}_k^{(t+1)} = v_k$, i.e. $\theta^{(t+1)}$ is the best $k$-sparse approximation of $v$, then we have the following from Lemma 3.3

$$\|\theta^{(t+1)} - \theta^*\|_2^2 \leq \left(1 + \frac{2\sqrt{k^*}}{\sqrt{k - k^*}}\right) \cdot \|v - \theta^*\|_2^2. \quad (7.2)$$

Let $\alpha = 1 + \frac{2\sqrt{k^*}}{\sqrt{k-k^*}}$. Combining (7.1) and (7.2), we have

$$\begin{aligned}
\mathbb{E}\|\theta^{(t+1)} - \theta^*\|_2^2 &\leq \alpha\mathbb{E}\|\theta^{(t)} - \theta^*\|_2^2 - 2\alpha\eta(1 - 6\eta\rho_s^+)\left[\mathcal{F}(\theta^{(t)}) - \mathcal{F}(\theta^*)\right] \\
&\quad + 12\alpha\eta^2 \rho_s^+ \left[\mathcal{F}(\widetilde{\theta}) - \mathcal{F}(\theta^*)\right] + 3\alpha\eta^2 \|\nabla_{\mathcal{I}}\mathcal{F}(\theta^*)\|_2^2. \quad (7.3)
\end{aligned}$$

Notice that $\widetilde{\theta} = \theta^{(0)} = \widetilde{\theta}^{(r-1)}$. By summing (7.3) over $t = 0, 1, \ldots, m-1$ and taking expectation with respect to all $t$'s, we have

$$\begin{aligned}
&\mathbb{E}\|\theta^{(m)} - \theta^*\|_2^2 + \frac{2\eta(1 - 6\eta\rho_s^+)(\alpha^m - 1)}{\alpha - 1}\mathbb{E}\left[\mathcal{F}(\widetilde{\theta}^{(r)}) - \mathcal{F}(\theta^*)\right] \\
&\leq \alpha^m \mathbb{E}\|\widetilde{\theta}^{(r-1)} - \theta^*\|_2^2 + \frac{12\eta^2 \rho_s^+(\alpha^m - 1)}{\alpha - 1}\mathbb{E}\left[\mathcal{F}(\widetilde{\theta}^{(r-1)}) - \mathcal{F}(\theta^*)\right] + \frac{3\eta^2(\alpha^m - 1)}{\alpha - 1}\mathbb{E}\|\nabla_{\mathcal{I}}\mathcal{F}(\theta^*)\|_2^2 \\
&\leq \frac{2\alpha^m}{\rho_s^-}\mathbb{E}\left[\mathcal{F}(\widetilde{\theta}^{(r-1)}) - \mathcal{F}(\theta^*) - \langle\nabla\mathcal{F}(\theta^*), \widetilde{\theta}^{(r-1)} - \theta^*\rangle\right] + \frac{12\eta^2 \rho_s^+(\alpha^m - 1)}{\alpha - 1}\mathbb{E}\left[\mathcal{F}(\widetilde{\theta}^{(r-1)}) - \mathcal{F}(\theta^*)\right] \\
&\quad + \frac{3\eta^2(\alpha^m - 1)}{\alpha - 1}\|\nabla_{\widetilde{\mathcal{I}}}\mathcal{F}(\theta^*)\|_2^2, \quad (7.4)
\end{aligned}$$

where the second inequality follows from the RSC condition (3.1) and the definition of $\widetilde{\mathcal{I}}$. It further follows from (7.4) that

$$\begin{aligned}
\mathbb{E}\left[\mathcal{F}(\widetilde{\theta}^{(r)}) - \mathcal{F}(\theta^*)\right] &\leq \left(\frac{\alpha^m(\alpha - 1)}{\eta\rho_s^-(1 - 6\eta\rho_s^+)(\alpha^m - 1)} + \frac{6\eta\rho_s^+}{1 - 6\eta\rho_s^+}\right)\mathbb{E}\left[\mathcal{F}(\widetilde{\theta}^{(r-1)}) - \mathcal{F}(\theta^*)\right] \\
&\quad + \frac{3\eta}{2(1 - 6\eta\rho_s^+)}\|\nabla_{\widetilde{\mathcal{I}}}\mathcal{F}(\theta^*)\|_2^2 + \frac{\alpha^m(\alpha - 1)}{\eta\rho_s^-(1 - 6\eta\rho_s^+)(\alpha^m - 1)}\left|\mathbb{E}\left[\langle\nabla\mathcal{F}(\theta^*), \widetilde{\theta}^{(r-1)} - \theta^*\rangle\right]\right|. \quad (7.5)
\end{aligned}$$

Let $\beta = \frac{\alpha^m(\alpha-1)}{\eta\rho_s^-(1-6\eta\rho_s^+)(\alpha^m-1)} + \frac{6\eta\rho_s^+}{1-6\eta\rho_s^+}$ and apply (7.5) recursively, then we have the desired bound (3.7) when $\beta \leq \frac{3}{4}$, $\frac{\alpha^m(\alpha-1)}{\eta\rho_s^-(1-6\eta\rho_s^+)(\alpha^m-1)} \leq \frac{1}{4}$, and

$$\mathbb{E}\langle\nabla\mathcal{F}(\theta^*), \theta^* - \widetilde{\theta}^{(r-1)}\rangle \leq \|\nabla\mathcal{F}(\theta^*)\|_\infty \mathbb{E}\|\widetilde{\theta}^{(r-1)} - \theta^*\|_1 \leq \sqrt{s}\|\nabla\mathcal{F}(\theta^*)\|_\infty \mathbb{E}\|\widetilde{\theta}^{(r-1)} - \theta^*\|_2.$$

We then demonstrate (3.8). The RSC condition implies

$$\mathcal{F}(\theta^*) \leq \mathcal{F}(\widetilde{\theta}^{(r)}) + \langle\nabla\mathcal{F}(\theta^*), \theta^* - \widetilde{\theta}^{(r)}\rangle - \frac{\rho_s^-}{2}\|\widetilde{\theta}^{(r)} - \theta^*\|_2^2. \tag{7.6}$$

Let $\zeta = \left(\frac{3}{4}\right)^r\left[\mathcal{F}(\widetilde{\theta}^{(0)}) - \mathcal{F}(\theta^*)\right] + \frac{6\eta}{(1-6\eta\rho_s^+)}\|\nabla_{\widetilde{\mathcal{I}}}\mathcal{F}(\theta^*)\|_2^2 + \sqrt{s}\|\nabla\mathcal{F}(\theta^*)\|_\infty \mathbb{E}\|\widetilde{\theta}^{(r-1)} - \theta^*\|_2$. Combining (3.7) and (7.6), we have

$$\mathbb{E}\left[\mathcal{F}(\widetilde{\theta}^{(r)}) - \zeta\right] \leq \mathcal{F}(\theta^*) \leq \mathbb{E}\left[\mathcal{F}(\widetilde{\theta}^{(r)}) + \langle\nabla\mathcal{F}(\theta^*), \theta^* - \widetilde{\theta}^{(r)}\rangle - \frac{\rho_s^-}{2}\|\widetilde{\theta}^{(r)} - \theta^*\|_2^2\right]. \tag{7.7}$$

In addition, we have

$$\mathbb{E}\langle\nabla\mathcal{F}(\theta^*), \theta^* - \widetilde{\theta}^{(r)}\rangle \leq \|\nabla\mathcal{F}(\theta^*)\|_\infty \mathbb{E}\|\widetilde{\theta}^{(r)} - \theta^*\|_1 \leq \sqrt{s}\|\nabla\mathcal{F}(\theta^*)\|_\infty \mathbb{E}\|\widetilde{\theta}^{(r)} - \theta^*\|_2. \tag{7.8}$$

Combining (7.7), (7.8), and $(\mathbb{E}[x])^2 \leq \mathbb{E}[x^2]$, we have

$$\frac{\rho_s^-}{2}(\mathbb{E}\|\widetilde{\theta}^{(r)} - \theta^*\|_2)^2 \leq \sqrt{s}\|\nabla\mathcal{F}(\theta^*)\|_\infty \mathbb{E}\|\widetilde{\theta}^{(r)} - \theta^*\|_2 + \zeta. \tag{7.9}$$

Let $a = \mathbb{E}\|\widetilde{\theta}^{(r)} - \theta^*\|_2$, then (7.9) is equivalent to solving the following quadratic function of $a$:

$$\frac{\rho_s^-}{2}a^2 - \sqrt{s}\|\nabla\mathcal{F}(\theta^*)\|_\infty a - \zeta \leq 0,$$

which yields the bound (3.8) from the solution of $a$ satisfying

$$a \leq \frac{\sqrt{s}\|\nabla\mathcal{F}(\theta^*)\|_\infty + \sqrt{2\rho_s^-\zeta}}{\rho_s^-} \leq \frac{\sqrt{s}\|\nabla\mathcal{F}(\theta^*)\|_\infty}{\rho_s^-} + \sqrt{\frac{2\left(\frac{3}{4}\right)^r\left[\mathcal{F}(\widetilde{\theta}^{(0)}) - \mathcal{F}(\theta^*)\right]}{\rho_s^-}}$$

$$+ \sqrt{\frac{12\eta}{\rho_s^-}(1-6\eta\rho_s^+)} \cdot \|\nabla_{\widetilde{\mathcal{I}}}\mathcal{F}(\theta^*)\|_2 + \sqrt{\frac{2\sqrt{s}\|\nabla\mathcal{F}(\theta^*)\|_\infty \mathbb{E}\|\widetilde{\theta}^{(r-1)} - \theta^*\|_2}{\rho_s^-}}. \tag{7.10}$$

Now we show that with $k$, $\eta$ and $m$ specified in the theorem, we guarantee $\beta \leq \frac{3}{4}$. More specifically, let $\eta \leq \frac{C_3}{\rho_s^+} \leq \frac{1}{18\rho_s^+}$, then we have

$$\frac{6\eta\rho_s^+}{1-6\eta\rho_s^+} \leq \frac{6C_3}{1-6C_3} \leq \frac{1}{2}.$$

If $k \geq C_1\kappa_s^2 k^*$ and $\eta \geq \frac{C_2}{\rho_s^+}$ with $C_2 \leq C_3$, then we have $\alpha \leq 1 + \frac{2}{\sqrt{C_1-1}\cdot\kappa_s}$ and

$$\frac{\alpha^m(\alpha-1)}{\eta\rho_s^-(1-6\eta\rho_s^+)(\alpha^m-1)} \leq \frac{\frac{2}{\sqrt{C_1-1}\cdot\kappa_s}}{\frac{2C_2}{3\kappa_s}\left(1-(1+\frac{2}{\sqrt{C_1-1}\cdot\kappa_s})^{-m}\right)} = \frac{3}{C_2\sqrt{C_1-1}\left(1-(1+\frac{2}{\sqrt{C_1-1}\cdot\kappa_s})^{-m}\right)}.$$

Then it is guaranteed $\frac{\alpha^m(\alpha-1)}{\eta\rho_s^-(1-6\eta\rho_s^+)(\alpha^m-1)} < \frac{1}{4}$ if we have

$$m \geq \log_{1+\frac{2}{\sqrt{C_1-1}\cdot\kappa_s}} \frac{C_2\sqrt{C_1-1}}{C_2\sqrt{C_1-1}-6}. \tag{7.11}$$

Using the the fact that $\ln(1+x) > x/2$ for $x \in (0,1)$, it follows that

$$\log_{1+\frac{2}{\sqrt{C_1-1}\cdot\kappa_s}} \frac{C_2\sqrt{C_1-1}}{C_2\sqrt{C_1-1}-6} = \frac{\log\frac{C_2\sqrt{C_1-1}}{C_2\sqrt{C_1-1}-6}}{\log 1 + \frac{2}{\sqrt{C_1-1}\cdot\kappa_s}} \leq \log\frac{C_2\sqrt{C_1-1}}{C_2\sqrt{C_1-1}-6} \cdot \sqrt{C_1-1}\cdot\kappa_s.$$

Then (7.11) holds if $m$ satisfies

$$m \geq \log \frac{C_2\sqrt{C_1-1}}{C_2\sqrt{C_1-1}-6} \cdot \sqrt{C_1-1}\cdot\kappa_s$$

If we choose $C_1 = 161^2$, $C_2 = \frac{1}{20}$, $C_3 = \frac{1}{18}$ and $C_4 = 222$, then we have $\beta \leq \frac{3}{4}$.

**Part 2.** Next, we demonstrate (3.10) and (3.11). It follows from (3.7)

$$\mathbb{E}\left[\mathcal{F}(\widetilde{\theta}^{(r)}) - \mathcal{F}(\theta^*)\right] - \frac{6\eta}{(1-6\eta\rho_s^+)}\|\nabla_{\widetilde{\mathcal{I}}}\mathcal{F}(\theta^*)\|_2^2 \leq \left(\frac{3}{4}\right)^r\left[\mathcal{F}(\widetilde{\theta}^{(0)}) - \mathcal{F}(\theta^*)\right]. \tag{7.12}$$

Let $\xi_1, \xi_2, \xi_3, \ldots$ be a non-negative sequence of random variables, which is defined as

$$\xi_r \triangleq \max\left\{\mathcal{F}(\widetilde{\theta}^{(r)}) - \mathcal{F}(\theta^*) - \frac{6\eta}{(1-6\eta\rho_s^+)}\|\nabla_{\widetilde{\mathcal{I}}}\mathcal{F}(\theta^*)\|_2^2,\ 0\right\}.$$

For a fixed $\varepsilon > 0$, it follows from Markov's Inequality and (7.12)

$$\mathbb{P}(\xi_r \geq \varepsilon) \leq \frac{\mathbb{E}\xi_r}{\varepsilon} \leq \frac{\left(\frac{3}{4}\right)^r\left[\mathcal{F}(\widetilde{\theta}^{(0)}) - \mathcal{F}(\theta^*)\right]}{\varepsilon}. \tag{7.13}$$

Given $\delta \in (0,1)$, let the R.H.S. of (7.13) be no greater than $\delta$, which requires

$$r \geq \log_{(\frac{4}{3})} \frac{\mathcal{F}(\widetilde{\theta}^{(0)}) - \mathcal{F}(\theta^*)}{\varepsilon\delta}.$$

Therefore, we have that if $r = \left\lceil 4\log\left(\frac{\mathcal{F}(\widetilde{\theta}^{(0)}) - \mathcal{F}(\theta^*)}{\varepsilon\delta}\right)\right\rceil$, then with probability at least $1-\delta$,

$$\left(\frac{3}{4}\right)^r\left[\mathcal{F}(\widetilde{\theta}^{(0)}) - \mathcal{F}(\theta^*)\right] \leq \varepsilon. \tag{7.14}$$

Using similar argument, if $r$ satisfies

$$r \geq \log_{(\frac{4}{3})} \frac{2\rho_s^-\left[\mathcal{F}(\widetilde{\theta}^{(0)}) - \mathcal{F}(\theta^*)\right]}{s\|\nabla\mathcal{F}(\theta^*)\|_\infty^2},$$

then we have

$$\sqrt{\frac{2\left(\frac{3}{4}\right)^r\left[\mathcal{F}(\widetilde{\theta}^{(0)}) - \mathcal{F}(\theta^*)\right]}{\rho_s^-}} \leq \frac{\sqrt{s}\|\nabla\mathcal{F}(\theta^*)\|_\infty}{\rho_s^-}.$$

Denote $b = \frac{\sqrt{s}\|\nabla \mathcal{F}(\theta^*)\|_\infty}{\rho_s^-}$. From our choice of $\eta$, we immediately have $\sqrt{\frac{12\eta}{\rho_s^-}(1-6\eta\rho_s^+)} \cdot \|\nabla_{\widetilde{\mathcal{I}}}\mathcal{F}(\theta^*)\|_2 \leq \frac{\sqrt{s}\|\nabla \mathcal{F}(\theta^*)\|_\infty}{\rho_s^-}$. Then it follows from (7.10) that

$$\mathbb{E}\|\widetilde{\theta}^{(r)} - \theta^*\|_2 \leq 3b + \sqrt{2b\mathbb{E}\|\widetilde{\theta}^{(r-1)} - \theta^*\|_2}.$$

Before further analysis, we state an intermediate result as following.

**Lemma 7.1.** Given a sequence $\{p_r\}$, $p_r > 0$ for all $r \in \mathbb{N}$, which satisfies

$$p_r \leq b + c\sqrt{bp_{r-1}},$$

where $b, c > 0$ are real constants. Denote $v = \frac{2+c^2+c\sqrt{c^2+4}}{2}$, then if $p_r \geq vb$, we have

$$\max\{p_r - vb, 0\} \leq \frac{c}{4\sqrt{v}}\max\{p_{r-1} - vb, 0\}.$$

Here we are only interested in $p_r = \mathcal{O}(b)$ since the R.H.S. corresponds to the optimal statistical error we address in the statistical analysis. Taking $p_r = \mathbb{E}\|\widetilde{\theta}^{(r)} - \theta^*\|_2$, we have from Lemma 7.1 that the sequence $\{p_r\}$ satisfies

$$\max\{p_r - vb, 0\} \leq \frac{1}{2}\max\{p_{r-1} - vb, 0\},$$

where $v = 4 + \sqrt{7}$ and $p_r \geq vb$. In other words, using analogous analysis above, given $\varepsilon > 0$, we have that if $r$ satisfies

$$r \geq \log_2 \frac{\left|\|\widetilde{\theta}^{(0)} - \theta^*\|_2 - vb\right|}{\varepsilon\delta},$$

then with probability at least $1 - \delta$, we have

$$\|\widetilde{\theta}^{(r)} - \theta^*\|_2 \leq \varepsilon.$$

Summarizing all result above, we have that if $r$ satisfies

$$r = \left\lceil \max\left\{4\log\left(\frac{\mathcal{F}(\widetilde{\theta}^{(0)}) - \mathcal{F}(\theta^*)}{\varepsilon\delta}\right), 4\log\frac{2\rho_s^-[\mathcal{F}(\widetilde{\theta}^{(0)}) - \mathcal{F}(\theta^*)]}{s\|\nabla\mathcal{F}(\theta^*)\|_\infty^2}\right\} + 2\log\frac{\left|\|\widetilde{\theta}^{(0)} - \theta^*\|_2 - vb\right|}{\varepsilon\delta}\right\rceil.$$

then (3.10) holds with probability at least $1 - 2\delta$ when $\varepsilon$ is as specified. Finally, (3.11) holds by combining (3.8) and (3.10).

## 7.2 Proof of Corollary 3.10

For sparse linear model, we have $\nabla\mathcal{F}(\theta^*) = A^\top z/(nb)$. Since $z$ has i.i.d. $\mathcal{N}(0, \sigma^2)$ entries, then $A_{*j}^\top z/(nb) \sim \mathcal{N}(0, \sigma^2 \|A_{*j}\|_2^2/(nb)^2)$ for any $j \in [d]$. Using Mill's Inequality for tail bounds of the normal distribution (Theorem 4.7 in Wasserman (2013)), we have

$$\mathbb{P}\left(\left|\frac{A_{*j}^\top z}{nb}\right| > 2\sigma\sqrt{\frac{\log d}{nb}}\right) = \mathbb{P}\left(\left|\frac{A_{*j}^\top z}{\sigma\|A_{*j}\|_2}\right| > 2\frac{\sqrt{nb\log d}}{\|A_{*j}\|_2}\right) \leq \|A_{*j}\|_2\sqrt{\frac{1}{2\pi nb\log d}}\exp\left(-4\frac{nb\log d}{\|A_{*j}\|_2^2}\right).$$

Using union bound and the assumption $\frac{\max_j \|A_{*j}\|_2}{\sqrt{nb}} \leq 1$, this implies

$$\mathbb{P}\left(\left\|\frac{A_{*j}^\top z}{nb}\right\|_\infty > 2\sigma\sqrt{\frac{\log d}{nb}}\right) \leq \frac{d^{-4}}{\sqrt{2\pi \log d}}.$$

Then with probability at least $1 - \frac{1}{\sqrt{2\pi \log d}} \cdot d^{-4}$, we have

$$\|\nabla \mathcal{F}(\theta^*)\|_\infty \leq \left\|\frac{A^\top z}{nb}\right\|_\infty \leq 2\sigma\sqrt{\frac{\log d}{nb}}. \tag{7.15}$$

Conditioning on (7.15), we have

$$\|\nabla_{\widetilde{\mathcal{I}}} \mathcal{F}(\theta^*)\|_2^2 \leq s\|\nabla \mathcal{F}(\theta^*)\|_\infty^2 \leq \frac{4\sigma^2 s \log d}{nb}. \tag{7.16}$$

We have from Lemma 3.9 that $s = 2k + k^* = (2C_5 + 1)k^*$ for some constant $C_5$ when $n$ and $b$ are large enough. Given $\varepsilon > 0$ and $\delta \in (0,1)$, if

$$r \geq 4\log\left(\frac{\mathcal{F}(\widetilde{\theta}^{(0)}) - \mathcal{F}(\theta^*)}{\varepsilon \delta}\right),$$

and $\varepsilon = \mathcal{O}\left(\sigma\sqrt{\frac{k^* \log d}{nb}}\right)$, then with probability at least $1 - \delta - \frac{1}{\sqrt{2\pi \log d}} \cdot d^{-4}$, we have from (3.11), (7.15), (7.16), and

$$\|\widetilde{\theta}^{(r)} - \theta^*\|_2 \leq c_3 \sigma \sqrt{\frac{k^* \log d}{nb}}, \tag{7.17}$$

where $c_3$ is a constant. This completes the proof.

## 7.3 Proof of Corollary 3.15

The only difference between the proof of Corollary 3.15 and the proof of Corollary 3.10 is the upper bounds of $\|\nabla \mathcal{F}(\theta^*)\|_\infty$ and $\|\nabla_{\widetilde{\mathcal{I}}} \mathcal{F}(\theta^*)\|_2^2$. When $\{A_{i*}\}_{i=1}^{nb}$ are independent sub-Gaussian vectors, it follows from Loh and Wainwright (2013) that $\mathcal{F}(\theta)$ and $\{f_i(\theta)\}_{i=1}^n$ satisfy (3.16). Besides, there exist constants $c_4, c_5$, and $c_6$, such that with probability at least $1 - c_4 d^{-c_5}$, we have

$$\|\nabla \mathcal{F}(\theta^*)\|_\infty \leq c_6 \sqrt{\frac{\log d}{nb}}. \tag{7.18}$$

Conditioning on (7.18), we have

$$\|\nabla_{\widetilde{\mathcal{I}}} \mathcal{F}(\theta^*)\|_2^2 \leq s\|\nabla \mathcal{F}(\theta^*)\|_\infty^2 \leq \frac{c_6^2 s \log d}{nb}. \tag{7.19}$$

The rest of the proof follows immediately from the proof of Corollary 3.10.

## 7.4 Proof of Theorem 4.1

Recall from (3.4) that $g^{(t)}(\theta^{(t)}) = \nabla f_{i_t}(\theta^{(t)}) - \nabla f_{i_t}(\widetilde{\theta}) + \nabla \mathcal{F}(\widetilde{\theta})$. We also denote $u = \theta^{(t)} - \eta h_{\mathcal{I}}^{(t)}(\theta^{(t)})$, where $h^{(t)}(\theta^{(t)}) = \nabla f_{i_t}(\theta^{(t')}) - \nabla f_{i_t}(\widetilde{\theta}) + \nabla \mathcal{F}(\widetilde{\theta})$ and $t'$ is the actual evaluation used at the $t$-th iteration. Then we have

$$\mathbb{E}\|u - \theta^*\|_2^2 = \mathbb{E}\|\theta^{(t)} - \eta h_{\mathcal{I}}^{(t)}(\theta^{(t)}) - \theta^*\|_2^2$$
$$= \mathbb{E}\left[\|\theta^{(t)} - \theta^*\|_2^2 + \eta^2\|h_{\mathcal{I}}^{(t)}(\theta^{(t)})\|_2^2 - 2\eta\langle\theta^{(t)} - \theta^*, h_{\mathcal{I}}^{(t)}(\theta^{(t)})\rangle\right] \quad (7.20)$$

We first bound $\mathbb{E}\|h_{\mathcal{I}}^{(t)}(\theta^{(t)})\|_2^2$ in terms of $\mathbb{E}\|g_{\mathcal{I}}^{(t)}(\theta^{(t)})\|_2^2$ as

$$\mathbb{E}\|h_{\mathcal{I}}^{(t)}(\theta^{(t)})\|_2^2 \leq 2\mathbb{E}\left[\|h_{\mathcal{I}}^{(t)}(\theta^{(t)}) - g_{\mathcal{I}}^{(t)}(\theta^{(t)})\|_2^2 + \|g_{\mathcal{I}}^{(t)}(\theta^{(t)})\|_2^2\right]$$
$$= 2\mathbb{E}\left[\|\nabla_{\mathcal{I}} f_{i_t}(\theta^{(t)}) - \nabla_{\mathcal{I}} f_{i_t}(\theta^{(t')})\|_2^2 + \|g_{\mathcal{I}}^{(t)}(\theta^{(t)})\|_2^2\right]$$
$$\leq 2(\rho_s^+)^2 \varsigma \sum_{j=t'}^{t-1} \mathbb{E}\|\theta_{e_t}^{(j+1)} - \theta_{e_t}^{(j)}\|^2 + 2\mathbb{E}\|g_{\mathcal{I}}^{(t)}(\theta^{(t)})\|_2^2$$
$$\leq 2(\rho_s^+)^2 \Delta \varsigma \eta^2 \sum_{j=t'}^{t-1} \mathbb{E}\|h_{\mathcal{I}}^{(j)}(\theta^{(j)})\|^2 + 2\mathbb{E}\|g_{\mathcal{I}}^{(t)}(\theta^{(t)})\|_2^2,$$

where the first inequality is from $\|a\|_2^2 \leq 2\|a-b\|_2^2 + 2\|b\|_2^2$ for any vector $a$ and $b$, the second inequality is from the definition of $\varsigma$, triangle inequality, and $\|f_i(\theta) - f_i(\theta')\|_2 \leq \rho_s^+\|\theta - \theta'\|_2$ implied by the RSS condition (Nesterov, 2013b), and the last inequality is from the definition of $\Delta$. Take the summation of the inequality above from $t = 0$ to $m-1$, we have

$$\sum_{t=0}^{m-1} \mathbb{E}\|h_{\mathcal{I}}^{(t)}(\theta^{(t)})\|_2^2 \leq \sum_{t=0}^{m-1}\left[2(\rho_s^+)^2 \Delta \varsigma \eta^2 \sum_{j=t'}^{t-1} \mathbb{E}\|h_{\mathcal{I}}^{(j)}(\theta^{(j)})\|^2 + 2\mathbb{E}\|g_{\mathcal{I}}^{(t)}(\theta^{(t)})\|_2^2\right]$$
$$\leq 2(\rho_s^+)^2 \Delta \varsigma^2 \eta^2 \sum_{t=0}^{m-1} \mathbb{E}\|h_{\mathcal{I}}^{(t)}(\theta^{(t)})\|_2^2 + \sum_{t=0}^{m-1} \mathbb{E}\|g_{\mathcal{I}}^{(t)}(\theta^{(t)})\|_2^2,$$

where the second inequality is from the definition of $\varsigma$. The inequality above implies

$$\sum_{t=0}^{m-1} \mathbb{E}\|h_{\mathcal{I}}^{(t)}(\theta^{(t)})\|_2^2 \leq \frac{2}{1 - 2\rho_s^{+2}\Delta\varsigma^2\eta^2} \sum_{t=0}^{m-1} \mathbb{E}\|g_{\mathcal{I}}^{(t)}(\theta^{(t)})\|_2^2. \quad (7.21)$$

Next, we bound $\mathbb{E}\langle\theta^{(t)} - \theta^*, h_{\mathcal{I}}^{(t)}(\theta^{(t)})\rangle$. This can be written as

$$\mathbb{E}\langle\theta^* - \theta^{(t)}, h_{\mathcal{I}}^{(t)}(\theta^{(t)})\rangle = \mathbb{E}\langle\theta^* - \theta^{(t)}, \nabla_{\mathcal{I}} f_{i_t}(\theta^{(t')})\rangle$$
$$= \mathbb{E}\langle\theta^* - \theta^{(t')}, \nabla_{\mathcal{I}} f_{i_t}(\theta^{(t')})\rangle + \sum_{j=t'}^{t-1} \mathbb{E}\langle\theta^{(j)} - \theta^{(j+1)}, \nabla_{\mathcal{I}} f_{i_t}(\theta^{(j)})\rangle$$
$$+ \sum_{j=t'}^{t-1} \mathbb{E}\langle\theta^{(j)} - \theta^{(j+1)}, \nabla_{\mathcal{I}} f_{i_t}(\theta^{(t')}) - \nabla_{\mathcal{I}} f_{i_t}(\theta^{(j)})\rangle. \quad (7.22)$$

From the restricted convexity of $f_{i_t}$, we have

$$\mathbb{E}\langle\theta^* - \theta^{(t')}, \nabla_{\mathcal{I}} f_{i_t}(\theta^{(t')})\rangle \leq \mathbb{E}\left[f_{i_t}(\theta^*) - f_{i_t}(\theta^{(t')})\right]. \tag{7.23}$$

Besides, the RSS condition implies

$$\sum_{j=t'}^{t-1} \mathbb{E}\langle\theta^{(j)} - \theta^{(j+1)}, \nabla_{\mathcal{I}} f_{i_j}(\theta^{(j)})\rangle \leq \sum_{j=t'}^{t-1} \mathbb{E}\left[f_{i_t}(\theta^{(j)}) - f_{i_t}(\theta^{(j+1)}) + \frac{\rho_s^+}{2}\|\theta^{(j)} - \theta^{(j+1)}\|_2^2\right]$$

$$\leq \mathbb{E}\left[f_{i_t}(\theta^{(t')}) - f_{i_t}(\theta^{(t)})\right] + \frac{\rho_s^+ \Delta}{2} \sum_{j=t'}^{t-1} \mathbb{E}\|\theta^{(j)} - \theta^{(j+1)}\|_2^2. \tag{7.24}$$

Moreover, we have

$$\sum_{j=t'}^{t-1} \mathbb{E}\langle\theta^{(j)} - \theta^{(j+1)}, \nabla_{\mathcal{I}} f_{i_t}(\theta^{(t')}) - \nabla_{\mathcal{I}} f_{i_t}(\theta^{(j)})\rangle$$

$$\leq \mathbb{E}\left[\sum_{j=t'}^{t-1} \|\theta_{e_t}^{(j)} - \theta_{e_t}^{(j+1)}\|_2 \cdot \|\nabla_{\mathcal{I}} f_{i_t}(\theta^{(t')}) - \nabla_{\mathcal{I}} f_{i_t}(\theta^{(j)})\|_2\right]$$

$$\leq \mathbb{E}\left[\sum_{j=t'}^{t-1} \|\theta_{e_t}^{(j)} - \theta_{e_t}^{(j+1)}\|_2 \cdot \sum_{l=t'}^{j-1} \|\nabla_{\mathcal{I}} f_{i_t}(\theta^{(l)}) - \nabla_{\mathcal{I}} f_{i_t}(\theta^{(l+1)})\|_2\right]$$

$$\leq \mathbb{E}\left[\sum_{j=t'}^{t-1} \sum_{l=t'}^{j-1} \frac{\rho_s^+}{2} \left(\|\theta_{e_t}^{(j)} - \theta_{e_t}^{(j+1)}\|_2 + \|\theta_{e_t}^{(l)} - \theta_{e_t}^{(l+1)}\|_2\right)\right]$$

$$\leq \frac{\rho_s^+ \Delta(\varsigma - 1)}{2} \sum_{j=t'}^{t-1} \mathbb{E}\|\theta^{(j)} - \theta^{(j+1)}\|_2^2, \tag{7.25}$$

where the first inequality is from Cauchy-Schwarz inequality, the second inequality is from the triangle inequality, the third inequality is from the RSS condition and the inequality of arithmetic and geometric means, and the last inequality is from a counting argument.

Combining (7.22) – (7.25), we have

$$\mathbb{E}\langle\theta^{(t)} - \theta^*, h_{\mathcal{I}}^{(t)}(\theta^{(t)})\rangle \geq \mathbb{E}\left[\mathcal{F}(\theta^{(t)} - \mathcal{F}(\theta^*) - \rho_s^+ \Delta\varsigma\eta^2 \sum_{j=t'}^{t-1} \mathbb{E}\|\theta^{(j)} - \theta^{(j+1)}\|_2^2\right]. \tag{7.26}$$

Combing (7.20), (7.21), and (7.26), we have

$$\mathbb{E}\|u - \theta^*\|_2^2 \leq \mathbb{E}\left[\|\theta^{(t)} - \theta^*\|_2^2 + \eta^2\|h_{\mathcal{I}}^{(t)}(\theta^{(t)})\|_2^2 - 2\eta\left(\mathcal{F}(\theta^{(t)} - \mathcal{F}(\theta^*)\right) + \rho_s^+ \Delta\varsigma\eta^2 \sum_{j=t'}^{t-1} \|h_{\mathcal{I}}^{(j)}(\theta^{(j)})\|_2^2\right]. \tag{7.27}$$

The rest of the proof follows analogously from the proof of Theorem 3.7. Specifically, by summing (7.27) over $t = 0, 1, \ldots, m-1$, taking expectation with respect to all $t$'s, and combining Lemma 3.3,

Lemma 3.5, and (7.26), we have

$$\mathbb{E}\|\theta^{(m)} - \theta^*\|_2^2 + \frac{2\eta(1 - 12\rho_s^+ \eta\Gamma)(\alpha^m - 1)}{\alpha - 1}\mathbb{E}\left[\mathcal{F}(\widetilde{\theta}^{(r)}) - \mathcal{F}(\theta^*)\right]$$
$$\leq \left(\frac{2\alpha^m}{\rho_s^-} + \frac{24\rho_s^+ \eta^2 \Gamma(\alpha^m - 1)}{\alpha - 1}\right)\mathbb{E}\left[\mathcal{F}(\widetilde{\theta}^{(r-1)}) - \mathcal{F}(\theta^*)\right] + \frac{6\eta^2 \Gamma(\alpha^m - 1)}{\alpha - 1}\|\nabla_{\widetilde{\mathcal{I}}}\mathcal{F}(\theta^*)\|_2^2$$
$$+ \frac{2\alpha^m}{\rho_s^-}\left|\mathbb{E}\langle \nabla\mathcal{F}(\theta^*), \widetilde{\theta}^{(r-1)} - \theta^*\rangle\right|, \tag{7.28}$$

where $\alpha = 1 + \frac{2\sqrt{k^*}}{\sqrt{k-k^*}}$ and $\Gamma = \frac{1+\rho_s^+ \Delta_\varsigma^2 \eta}{1-2\rho_s^{+2}\Delta_\varsigma^2 \eta^2}$. It further follows from (7.28)

$$\mathbb{E}\left[\mathcal{F}(\widetilde{\theta}^{(r)}) - \mathcal{F}(\theta^*)\right] \leq \left(\frac{\alpha^m(\alpha - 1)}{\eta\rho_s^-(1 - 12\eta\rho_s^+ \Gamma)(\alpha^m - 1)} + \frac{12\eta\rho_s^+ \Gamma}{1 - 12\eta\rho_s^+ \Gamma}\right)\mathbb{E}\left[\mathcal{F}(\widetilde{\theta}^{(r-1)}) - \mathcal{F}(\theta^*)\right]$$
$$+ \frac{3\eta\Gamma}{1 - 12\eta\rho_s^+ \Gamma}\|\nabla_{\widetilde{\mathcal{I}}}\mathcal{F}(\theta^*)\|_2^2 + \frac{\alpha^m(\alpha - 1)}{\eta\rho_s^-(1 - 12\eta\rho_s^+)(\alpha^m - 1)}\left|\mathbb{E}\left[\langle \nabla\mathcal{F}(\theta^*), \widetilde{\theta}^{(r-1)} - \theta^*\rangle\right]\right|. \tag{7.29}$$

Finally, $\frac{\alpha^m(\alpha-1)}{\eta\rho_s^-(1-12\eta\rho_s^+ \Gamma)(\alpha^m-1)} + \frac{12\eta\rho_s^+ \Gamma}{1-12\eta\rho_s^+ \Gamma} \leq \frac{5}{6}$ holds with the same choices of constants $C_1$ to $C_4$ as in Theorem 3.7 and $C_5 = \frac{1}{2}$.

# 8 Acknowledgment

We thank Quanquan Gu for helpful discussions on the proof techniques.

# 9 Appendix

## 9.1 Further Examples of Nonconvex Loss $\mathcal{F}$

We provide further discussion where $A$ has additive noise or multiplicative noise.

(1) **Additive noise**. Suppose that we observe

$$Z = A + W,$$

where $W$ is a random matrix with i.i.d. rows drawn from a zero mean distribution that has a known covariance matrix $\Sigma_W$, independent of $A$. Then we set

$$\widehat{\Gamma} = \frac{Z^\top Z}{nb} - \Sigma_W \text{ and } \widehat{b} = \frac{Z^\top y}{nb}.$$

For $\Sigma_W \neq 0$, since $\frac{Z^\top Z}{nb}$ has rank at most of $nb$, the subtraction of matrix $\Sigma_W$ may cause $\widehat{\Gamma}$ to have negative eigenvalues when $d \gg nb$, hence a non-PSD $\widehat{\Gamma}$.

(2) **Multiplicative noise**. Suppose that we observe

$$Z = A \odot U,$$

where $\odot$ is the Hadamard (entry-wise) product and $U$ is a noise matrix with nonnegative entries, e.g., rows $u_i$ of $U$ are i.i.d. random vectors drawn from a distribution in which both $\mathbb{E}(u_i)$ and $\mathbb{E}(u_i^\top u_i)$ have strictly positive entries. Then we set

$$\widehat{\Gamma} = \frac{Z^\top Z}{nb} \oslash \mathbb{E}(u_i^\top u_i) \text{ and } \widehat{b} = \frac{Z^\top y}{nb} \oslash \mathbb{E}(u_i),$$

where $\oslash$ is entry-wise division. $\widehat{\Gamma}$ may have negative eigenvalues, hence a non-PSD $\widehat{\Gamma}$, when $d \gg nb$ since $\frac{Z^\top Z}{nb}$ has rank at most $nb$. This can also be viewed as a generalization of the missing data scenario in Remark 3.12, if entries $\{u_{ij}\}$ of $U$ are independent Bernoulli$(1-\rho)$ random variables.

## 9.2 Proof of Lemma 3.3

For notational convenience, denote $\theta' = \mathcal{H}_k(\theta)$. Let $\text{supp}(\theta^*) = \mathcal{I}^*$, $\text{supp}(\theta) = \mathcal{I}$, $\text{supp}(\theta') = \mathcal{I}'$, and $\theta'' = \theta - \theta'$ with $\text{supp}(\theta'') = \mathcal{I}''$. Clearly we have $\mathcal{I}' \cup \mathcal{I}'' = \mathcal{I}$, $\mathcal{I}' \cap \mathcal{I}'' = \emptyset$, and $\|\theta\|_2^2 = \|\theta'\|_2^2 + \|\theta''\|_2^2$. Then we have

$$\|\theta' - \theta^*\|_2^2 - \|\theta - \theta^*\|_2^2 = \|\theta'\|_2^2 - 2\langle \theta', \theta^* \rangle - \|\theta\|_2^2 + 2\langle \theta, \theta^* \rangle = 2\langle \theta'', \theta^* \rangle - \|\theta''\|_2^2. \tag{9.1}$$

If $2\langle \theta'', \theta^* \rangle - \|\theta''\|_2^2 \leq 0$, then (3.3) holds naturally. From this point on, we will discuss the situation when $2\langle \theta'', \theta^* \rangle - \|\theta''\|_2^2 > 0$.

Let $\mathcal{I}^* \cap \mathcal{I}' = \mathcal{I}^{*1}$ and $\mathcal{I}^* \cap \mathcal{I}'' = \mathcal{I}^{*2}$, and denote $(\theta^*)_{\mathcal{I}^{*1}} = \theta^{*1}$, $(\theta^*)_{\mathcal{I}^{*2}} = \theta^{*2}$, $(\theta')_{\mathcal{I}^{*1}} = \theta^{1*}$, and $(\theta'')_{\mathcal{I}^{*2}} = \theta^{2*}$. Then we have

$$2\langle \theta'', \theta^* \rangle - \|\theta''\|_2^2 = 2\langle \theta^{2*}, \theta^{*2} \rangle - \|\theta''\|_2^2 \leq 2\langle \theta^{2*}, \theta^{*2} \rangle - \|\theta^{2*}\|_2^2 \leq 2\|\theta^{2*}\|_2 \|\theta^{*2}\|_2 - \|\theta^{2*}\|_2^2. \tag{9.2}$$

Let $|\mathrm{supp}(\theta^{2*})| = |\mathcal{I}^{*2}| = k^{**}$ and $\theta_{2,\max} = \|\theta^{2*}\|_\infty$, then consequently we have $\|\theta^{2*}\|_2 = m \cdot \theta_{2,\max}$ for some $m \in [1, \sqrt{k^{**}}]$. Notice that we are interested in $1 \leq k^{**} \leq k^*$, since (3.3) holds naturally if $k^{**} = 0$. In terms of $\|\theta^{*2}\|_2$, the R.H.S. of (9.2) is maximized in the following three cases.

Case 1: $m = 1$, if $\|\theta^{*2}\|_2 \leq \theta_{2,\max}$;

Case 2: $m = \frac{\|\theta^{*2}\|_2}{\theta_{2,\max}}$, if $\theta_{2,\max} < \|\theta^{*2}\|_2 < \sqrt{k^{**}}\theta_{2,\max}$;

Case 3: $m = \sqrt{k^{**}}$, if $\|\theta^{*2}\|_2 \geq \sqrt{k^{**}}\theta_{2,\max}$.

Case 1. If $\|\theta^{*2}\|_2 \leq \theta_{2,\max}$, then the R.H.S. of (9.2) is maximized when $m = 1$, i.e. $\theta^{2*}$ has only one nonzero element $\theta_{2,\max}$. From (9.2), we have

$$2\langle \theta'', \theta^* \rangle - \|\theta''\|_2^2 \leq 2\theta_{2,\max}\|\theta^{*2}\|_2 - \theta_{2,\max}^2 \leq 2\theta_{2,\max}^2 - \theta_{2,\max}^2 = \theta_{2,\max}^2. \tag{9.3}$$

Denote $\theta_{1,\min}$ as the smallest element of $\theta^{1*}$ (in magnitude), which indicates that $|\theta_{1,\min}| \geq |\theta_{2,\max}|$ as $\theta'$ contains the largest $k$ entries and $\theta''$ contains the smallest $d - k$ entries of $\theta$. For $\|\theta - \theta^*\|_2^2$, we have

$$\|\theta - \theta^*\|_2^2 = \|\theta' - \theta^{*1}\|_2^2 + \|\theta'' - \theta^{*2}\|_2^2$$
$$= \|\theta_{(\mathcal{I}^{*1})^C}\|_2^2 + \|\theta_{\mathcal{I}^{*1}} - \theta^{*1}\|_2^2 + \|\theta^{*2}\|_2^2 - (2\langle \theta'', \theta^* \rangle - \|\theta''\|_2^2) \tag{9.4}$$
$$\geq (k - k^* + k^{**})\theta_{1,\min}^2 - \theta_{2,\max}^2, \tag{9.5}$$

where the last inequality follows from the fact that $\theta_{(\mathcal{I}^{*1})^C}$ has $k - k^* + k^{**}$ entries larger than $\theta_{1,\min}$ (in magnitude). Combining (9.1), (9.3), and (9.5), we have

$$\frac{\|\theta' - \theta^*\|_2^2 - \|\theta - \theta^*\|_2^2}{\|\theta - \theta^*\|_2^2} \leq \frac{\theta_{2,\max}^2}{(k - k^* + k^{**})\theta_{1,\min}^2 - \theta_{2,\max}^2}$$
$$\leq \frac{\theta_{2,\max}^2}{(k - k^* + k^{**})\theta_{2,\max}^2 - \theta_{2,\max}^2} \leq \frac{1}{k - k^*}. \tag{9.6}$$

Case 2. If $\theta_{2,\max} < \|\theta^{*2}\|_2 < \sqrt{k^{**}}\theta_{2,\max}$, then the R.H.S. of (9.2) is maximized when $m = \frac{\|\theta^{*2}\|_2}{\theta_{2,\max}}$. From (9.2), we have

$$2\langle \theta'', \theta^* \rangle - \|\theta''\|_2^2 \leq 2\sqrt{k^{**}}\theta_{2,\max} \cdot m\theta_{2,\max} - \theta_{2,\max}^2 \leq k^{**}\theta_{2,\max}^2. \tag{9.7}$$

From (9.4), we have

$$\|\theta - \theta^*\|_2^2 \geq (k - k^* + k^{**})\theta_{1,\min}^2 + m^2\theta_{2,\max}^2 - \theta_{2,\max}^2 \geq (k - k^* + k^{**})\theta_{1,\min}^2. \tag{9.8}$$

Combining (9.1), (9.7), and (9.8), we have

$$\frac{\|\theta' - \theta^*\|_2^2 - \|\theta - \theta^*\|_2^2}{\|\theta - \theta^*\|_2^2} \leq \frac{k^{**}\theta_{2,\max}^2}{(k - k^* + k^{**})\theta_{1,\min}^2} \leq \frac{k^{**}}{k - k^* + k^{**}}. \tag{9.9}$$

Case 3. If $\|\theta^{*2}\|_2 \geq \sqrt{k^{**}}\theta_{2,\max}$, then the R.H.S. of (9.2) is maximized when $m = \sqrt{k^{**}}$. Let $\|\theta^{*2}\|_2 = \gamma\theta_{2,\max}$ for some $\gamma \geq \sqrt{k^{**}}$. From (9.2), we have

$$2\langle \theta'', \theta^* \rangle - \|\theta''\|_2^2 \leq 2\gamma\sqrt{k^{**}}\theta_{2,\max}^2 - k^{**}\theta_{2,\max}^2. \tag{9.10}$$

From (9.4), we have

$$\|\theta - \theta^*\|_2^2 \geq (k - k^* + k^{**})\theta_{1,\min}^2 + \gamma^2\theta_{2,\max}^2 - \gamma\sqrt{k^{**}}\theta_{2,\max}^2 + k^{**}\theta_{2,\max}^2. \tag{9.11}$$

Combining (9.1), (9.10), and (9.11), we have

$$\frac{\|\theta' - \theta^*\|_2^2 - \|\theta - \theta^*\|_2^2}{\|\theta - \theta^*\|_2^2} \leq \frac{2\gamma\sqrt{k^{**}}\theta_{2,\max}^2 - k^{**}\theta_{2,\max}^2}{(k - k^* + k^{**})\theta_{1,\min}^2 + \gamma^2\theta_{2,\max}^2 - \gamma\sqrt{k^{**}}\theta_{2,\max}^2 + k^{**}\theta_{2,\max}^2}$$

$$\leq \frac{2\gamma\sqrt{k^{**}} - k^{**}}{k - k^* + 2k^{**} + \gamma^2 - 2\gamma\sqrt{k^{**}}}. \tag{9.12}$$

Inspecting the R.H.S. of (9.12) carefully, we can see that it is either a bell shape function or a monotone decreasing function when $\gamma \geq \sqrt{k^{**}}$. Setting the first derivative of the R.H.S. in terms of $\gamma$ to zero, we have $\gamma = \frac{1}{2}\sqrt{k^{**}} + \sqrt{k - k^* + \frac{5}{4}k^{**}}$ (the other root is smaller than $\sqrt{k^{**}}$). Denoting $\gamma_* = \max\{\sqrt{k^{**}}, \frac{1}{2}\sqrt{k^{**}} + \sqrt{k - k^* + \frac{5}{4}k^{**}}\}$ and plugging it into the R.H.S. of (9.12), we have

$$\frac{\|\theta' - \theta^*\|_2^2 - \|\theta - \theta^*\|_2^2}{\|\theta - \theta^*\|_2^2} \leq \max\left\{\frac{k^{**}}{k - k^* + k^{**}}, \frac{2\sqrt{k^{**}}}{2\sqrt{k - k^* + \frac{5}{4}k^{**}} - \sqrt{k^{**}}}\right\}. \tag{9.13}$$

Combining (9.6), (9.9), and (9.13), and taking $k > k^*$ and $k^* \geq k^{**} \geq 1$ into consideration, we have

$$\max\left\{\frac{1}{k - k^*}, \frac{k^{**}}{k - k^* + k^{**}}, \frac{2\sqrt{k^{**}}}{2\sqrt{k - k^* + \frac{5}{4}k^{**}} - \sqrt{k^{**}}}\right\} \leq \frac{2\sqrt{k^{**}}}{2\sqrt{k - k^* + \frac{5}{4}k^{**}} - \sqrt{k^{**}}}$$

$$\leq \frac{2\sqrt{k^*}}{2\sqrt{k - k^*} - \sqrt{k^*}} \leq \frac{2\sqrt{k^*}}{\sqrt{k - k^*}},$$

which finishes the proof.

## 9.3 Proof of Lemma 3.5

It is straightforward that the stochastic variance reduced gradient (3.4) satisfies

$$\mathbb{E}g^{(t)}(\theta^{(t)}) = \mathbb{E}\nabla f_{i_t}(\theta^{(t)}) - \mathbb{E}\nabla f_{i_t}(\widetilde{\theta}) + \widetilde{\mu} = \nabla \mathcal{F}(\theta^{(t)}).$$

Thus $g^{(t)}(\theta^{(t)})$ is a unbiased estimator of $\nabla \mathcal{F}(\theta^{(t)})$ and the first claim is verified.

For convenience, we ignore the superscript $(t)$ in the following analysis. Next, we bound $\mathbb{E}\|g_\mathcal{I}(\theta)\|_2^2$. For any $i \in [n]$ and $\theta$ with $\text{supp}(\theta) \subseteq \mathcal{I}$, consider

$$\phi_i(\theta) = f_i(\theta) - f_i(\theta^*) - \langle \nabla f_i(\theta^*), \theta - \theta^* \rangle.$$

Since $\nabla \phi_i(\theta^*) = \nabla f_i(\theta^*) - \nabla f_i(\theta^*) = 0$, we have $\phi_i(\theta^*) = \min_\theta \phi_i(\theta)$, which implies

$$0 = \phi_i(\theta^*) \leq \min_\eta \phi_i(\theta - \eta \nabla_\mathcal{I} \phi_i(\theta)) \leq \min_\eta \phi_i(\theta) - \eta \|\nabla_\mathcal{I} \phi_i(\theta)\|_2^2 + \frac{\rho_s^+ \eta^2}{2} \|\nabla_\mathcal{I} \phi_i(\theta)\|_2^2$$
$$= \phi_i(\theta) - \frac{1}{2\rho_s^+} \|\nabla_\mathcal{I} \phi_i(\theta)\|_2^2, \tag{9.14}$$

where the second inequality follows from the RSS condition and the last equality follows from the fact that $\eta = 1/\rho_s^+$ minimizes the function. From (9.14), we have

$$\|\nabla_\mathcal{I} f_i(\theta) - \nabla_\mathcal{I} f_i(\theta^*)\|_2^2 \leq 2\rho_s^+ [f_i(\theta) - f_i(\theta^*) - \langle \nabla_\mathcal{I} f_i(\theta^*), \theta - \theta^* \rangle]. \tag{9.15}$$

Since the sampling of $i$ from $[n]$ is uniform, we have from (9.15)

$$\mathbb{E}\|\nabla_\mathcal{I} f_i(\theta) - \nabla_\mathcal{I} f_i(\theta^*)\|_2^2 = \frac{1}{n} \sum_{i=1}^n \|\nabla_\mathcal{I} f_i(\theta) - \nabla_\mathcal{I} f_i(\theta^*)\|_2^2 \leq 2\rho_s^+ [\mathcal{F}(\theta) - \mathcal{F}(\theta^*) - \langle \nabla_\mathcal{I} \mathcal{F}(\theta^*), \theta - \theta^* \rangle]$$
$$\leq 2\rho_s^+ [\mathcal{F}(\theta) - \mathcal{F}(\theta^*) + |\langle \nabla_\mathcal{I} \mathcal{F}(\theta^*), \theta - \theta^* \rangle|] \leq 4\rho_s^+ [\mathcal{F}(\theta) - \mathcal{F}(\theta^*)], \tag{9.16}$$

where the last inequality is from the restricted convexity of $\mathcal{F}(\theta)$ and the fact that $\|(\theta - \theta^*)_{\mathcal{I}^c}\|_0 = 0$.

By the definition of $g_\mathcal{I}$ in (3.4), we can verify the second claim as

$$\mathbb{E}\|g_\mathcal{I}(\theta)\|_2^2 \leq 3\mathbb{E}\|\left[\nabla_\mathcal{I} f_{i_t}(\widetilde{\theta}) - \nabla_\mathcal{I} f_{i_t}(\theta^*)\right] - \nabla_\mathcal{I} \mathcal{F}(\widetilde{\theta}) + \nabla_\mathcal{I} \mathcal{F}(\theta^*)\|_2^2$$
$$+ 3\mathbb{E}\|\nabla_\mathcal{I} f_{i_t}(\theta) - \nabla_\mathcal{I} f_{i_t}(\theta^*)\|_2^2 + 3\|\nabla_\mathcal{I} \mathcal{F}(\theta^*)\|_2^2$$
$$\leq 3\mathbb{E}\|\nabla_\mathcal{I} f_{i_t}(\theta) - \nabla_\mathcal{I} f_{i_t}(\theta^*)\|_2^2 + 3\mathbb{E}\|\nabla_\mathcal{I} f_{i_t}(\widetilde{\theta}) - \nabla_\mathcal{I} f_{i_t}(\theta^*)\|_2^2 + 3\|\nabla_\mathcal{I} \mathcal{F}(\theta^*)\|_2^2$$
$$\leq 12\rho_s^+ \left[\mathcal{F}(\theta) - \mathcal{F}(\theta^*) + \mathcal{F}(\widetilde{\theta}) - \mathcal{F}(\theta^*)\right] + 3\|\nabla_\mathcal{I} \mathcal{F}(\theta^*)\|_2^2, \tag{9.17}$$

where the first inequality follows from the power mean inequality $\|a+b+c\|_2^2 \leq 3\|a\|_2^2 + 3\|b\|_2^2 + 3\|c\|_2^2$, the second inequality follows from $\mathbb{E}\|x - \mathbb{E}x\|_2^2 \leq \mathbb{E}\|x\|_2^2$ with $\mathbb{E}\left[\nabla_\mathcal{I} f_{i_t}(\widetilde{\theta}) - \nabla_\mathcal{I} f_{i_t}(\theta^*)\right] = \nabla_\mathcal{I} \mathcal{F}(\widetilde{\theta}) - \nabla_\mathcal{I} \mathcal{F}(\theta^*)$, and the last inequality follows from (9.16).

## 9.4 Proof of Lemma 3.9

For any $\theta, \theta' \in \mathbb{R}^d$ in sparse linear model, we have $\nabla^2 \mathcal{F}(\theta) = A^\top A$ and there exists some $\theta''$ such that

$$\mathcal{F}(\theta) - \mathcal{F}(\theta') - \langle \nabla \mathcal{F}(\theta'), \theta - \theta' \rangle = \frac{1}{2}(\theta - \theta')^\top \nabla^2 \mathcal{F}(\theta'')(\theta - \theta') = \frac{1}{2}\|A(\theta - \theta')\|_2^2,$$

where $\|\theta - \theta'\|_0 \leq 2k \leq s$. Let $v = \theta - \theta'$, then $\|v\|_0 \leq s$ and $\|v\|_1^2 \leq s\|v\|_2^2$. From (3.12), we have

$$\frac{\|Av\|_2^2}{nb} \geq \psi_1 \|v\|_2^2 - \varphi_1 \frac{s \log d}{nb} \|v\|_2^2 \text{ and } \frac{\|A_{\mathcal{S}_{i^*}} v\|_2^2}{b} \leq \psi_2 \|v\|_2^2 + \varphi_2 \frac{s \log d}{b} \|v\|_2^2, \forall i \in [n].$$

The inequality above further imply

$$\rho_s^- = \inf_{\|v\|_0 \leq s} \frac{\|Av\|_2^2}{nb\|v\|_2^2} \geq \psi_1 - \varphi_1 \frac{s \log d}{nb} \quad \text{and} \quad \rho_s^+ = \sup_{\|v\|_0 \leq s, i \in [n]} \frac{\|A_{S_i*}v\|_2^2}{b\|v\|_2^2} \leq \psi_2 + \varphi_2 \frac{s \log d}{b}. \quad (9.18)$$

If $b \geq \frac{\varphi_2 s \log d}{\psi_2}$ and $n \geq \frac{2\varphi_1 \psi_2}{\psi_1 \varphi_2}$, then we have $nb \geq \frac{2\varphi_1 s \log d}{\psi_1}$. Combining these with (9.18), we have

$$\rho_s^- \geq \frac{1}{2}\psi_1, \text{ and } \rho_s^+ \leq 2\psi_2.$$

This implies $\kappa_s = \frac{\rho_s^+}{\rho_s^-} \leq \frac{4\psi_2}{\psi_1}$. Then there exists some $C_5 \geq \frac{16 C_1 \psi_2^2}{\psi_1^2}$ such that

$$k = C_5 k^* \geq C_1 \kappa_s^2 k^*.$$

## 9.5 Proof of Lemma 3.19

Let $\Theta = U\Sigma V^\top$ and $\Theta^* = U^*\Sigma^* V^{*\top}$ be the singular value decomposition of $\Theta$ and $\Theta^*$ respectively, where $\Sigma$ and $\Sigma^*$ are . Since $\Sigma$ and $\Sigma^*$ are diagonal, if $k > k^*$, we have from Lemma 3.3

$$\|\mathcal{R}_k(\Sigma) - \Sigma^*\|_F^2 \leq \left(1 + \frac{2\sqrt{k^*}}{\sqrt{k - k^*}}\right) \|\Sigma - \Sigma^*\|_F^2. \quad (9.19)$$

Then we have

$$\|\mathcal{R}_k(\Theta) - \Theta^*\|_F^2 - \|\Theta - \Theta^*\|_F^2 = \|\mathcal{R}_k(\Theta)\|_F^2 - \|\Theta\|_F^2 + 2\langle \Theta - \mathcal{R}_k(\Theta), \Theta^* \rangle$$

$$= \|\mathcal{R}_k(\Sigma)\|_F^2 - \|\Sigma\|_F^2 + 2\langle \Theta - \mathcal{R}_k(\Theta), \Theta^* \rangle \leq \|\mathcal{R}_k(\Sigma)\|_F^2 - \|\Sigma\|_F^2 + 2\sum_{i=1}^{k^*} \sigma_i(\Theta - \mathcal{R}_k(\Theta)) \cdot \sigma_i(\Theta^*)$$

$$= \|\mathcal{R}_k(\Sigma)\|_F^2 - \|\Sigma\|_F^2 + 2\sum_{i=1}^{k^*} (\sigma_{i+k}(\Theta) - \sigma_{i+k}(\mathcal{R}_k(\Theta))) \cdot \sigma_i(\Theta^*) = \|\mathcal{R}_k(\Sigma) - \Sigma^*\|_F^2 - \|\Sigma - \Sigma^*\|_F^2$$

$$\leq \frac{2\sqrt{k^*}}{\sqrt{k - k^*}} \cdot \|\Sigma - \Sigma^*\|_F^2 \leq \frac{2\sqrt{k^*}}{\sqrt{k - k^*}} \cdot \|\Theta - \Theta^*\|_F^2,$$

where the first and last inequalities are from Von Neumann's trace inequality Mirsky (1975) $\langle A, B \rangle \leq \sum_{i=1}^{\min\{\text{rank}(A), \text{rank}(B)\}} \sigma_i(A) \cdot \sigma_i(B)$ for matrices $A, B \in \mathbb{R}^{d \times p}$, the second inequality is from (9.19), and the last inequality holds by rearranging $\Sigma^*$ such that the diagonal elements from $k + 1$ to $k + k^*$ are nonzero. This finishes the proof.

## 9.6 Proof of Lemma 7.1

Given $p_r \leq b + c\sqrt{bp_{r-1}}$ and some real $\nu > 0$, we have

$$p_r - \nu b \leq (1 - \nu)b + c\sqrt{bp_{r-1}}. \quad (9.20)$$

Take the Taylor expansion of the R.H.S. of $\sqrt{bp_{r-1}}$ with respect to $p_{r-1}$, we have

$$\sqrt{bp_{r-1}} \leq \sqrt{\nu}b + \frac{p_{r-1} - \nu b}{4\sqrt{\nu}}. \quad (9.21)$$

Combining (9.20) and (9.21), we have

$$p_r - \nu b \le (1-\nu)b + bc\sqrt{\nu} + \frac{c}{4\sqrt{\nu}}(p_{r-1} - \nu b). \tag{9.22}$$

Solving $(1-\nu)b + bc\sqrt{\nu} = 0$, we have $\nu = \frac{2+c^2+c\sqrt{c^2+4}}{2}$, then it follows from (9.22) that

$$p_r - \nu b \le \frac{c}{4\sqrt{\nu}}(p_{r-1} - \nu b),$$

which finishes the proof.